\documentclass[sigconf]{acmart}

\AtBeginDocument{%
  }

\copyrightyear{2026}
\acmYear{2026}
\setcopyright{rightsretained}
\acmDOI{10.1145/3675094.3678379}

\usepackage{booktabs}
\usepackage{multirow}
\usepackage{amsmath,amsfonts}
\usepackage{algorithmic}
\usepackage{algorithm}
\usepackage{array}
\usepackage[caption=false,font=normalsize,labelfont=sf,textfont=sf]{subfig}
\usepackage{textcomp}
\usepackage{stfloats}
\usepackage{url}
\usepackage{verbatim}
\usepackage{graphicx}
\usepackage{hhline}
\usepackage{makecell}
\usepackage{enumitem}
\usepackage{placeins}

\acmConference[CHI '26]{CHI}{April 13--17,
  2026}{Barcelona, Spain}
\acmBooktitle{CHI (CHI '26),
  April 13--17, 2026, Barcelona, Spain}
\acmISBN{0000}

\begin{document}

\title{PREFAB: PREFerence-based Affective Modeling for Low-Budget Self-Annotation}

\author{Jaeyoung Moon}
\orcid{0000-0002-1852-2769}
\affiliation{%
  \institution{Department of AI Convergence}
  \institution{Gwangju Institute of Science and Technology}
  \city{Gwangju}
  \country{South Korea}
}
\email{super\_moon@gm.gist.ac.kr}

\author{Youjin Choi}
\orcid{0000-0003-2788-7871}
\affiliation{%
  \institution{Department of AI Convergence}
  \institution{Gwangju Institute of Science and Technology}
  \city{Gwangju}
  \country{South Korea}
}
\email{chldbwls304@gm.gist.ac.kr}

\author{Yucheon Park}
\orcid{0009-0009-4630-4089}
\affiliation{%
  \institution{Department of AI Convergence}
  \institution{Gwangju Institute of Science and Technology}
  \city{Gwangju}
  \country{South Korea}
}
\email{yucheon6000@gm.gist.ac.kr}

\author{David Melhart}
\orcid{0000-0002-2692-7061}
\affiliation{%
  \institution{Metaverse Lab}
  \institution{University of Southern Denmark}
  \city{Odense}
  \country{Denmark}
}
\email{davm@mmmi.sdu.dk}

\author{Georgios N. Yannakakis}
\orcid{0000-0001-7793-1450}
\affiliation{%
  \institution{Institute of Digital Games}
  \institution{University of Malta}
  \city{Msida}
  \country{Malta}
}
\email{georgios.yannakakis@um.edu.mt}

\author{Kyung-Joong Kim}
\orcid{0000-0002-7732-0817}
\affiliation{%
  \institution{Department of AI Convergence}
  \institution{Gwangju Institute of Science and Technology}
  \city{Gwangju}
  \country{South Korea}
}
\email{kjkim@gist.ac.kr}
\renewcommand{\shortauthors}{Moon et al.}


\begin{abstract}
Self-annotation is the gold standard for collecting affective state labels in affective computing. Existing methods typically rely on full annotation, requiring users to continuously label affective states across entire sessions. While this process yields fine-grained data, it is time-consuming, cognitively demanding, and prone to fatigue and errors. To address these issues, we present PREFAB, a low-budget retrospective self-annotation method that targets affective inflection regions rather than full annotation. Grounded in the peak-end rule and ordinal representations of emotion, PREFAB employs a preference-learning model to detect relative affective changes, directing annotators to label only selected segments while interpolating the remainder of the stimulus. We further introduce a preview mechanism that provides brief contextual cues to assist annotation. We evaluate PREFAB through a technical performance study and a 25-participant user study. Results show that PREFAB outperforms baselines in modeling affective inflections while mitigating workload (and conditionally mitigating temporal burden). Importantly PREFAB improves annotator confidence without degrading annotation quality.
\end{abstract}


\begin{teaserfigure}
  \includegraphics[width=\columnwidth]{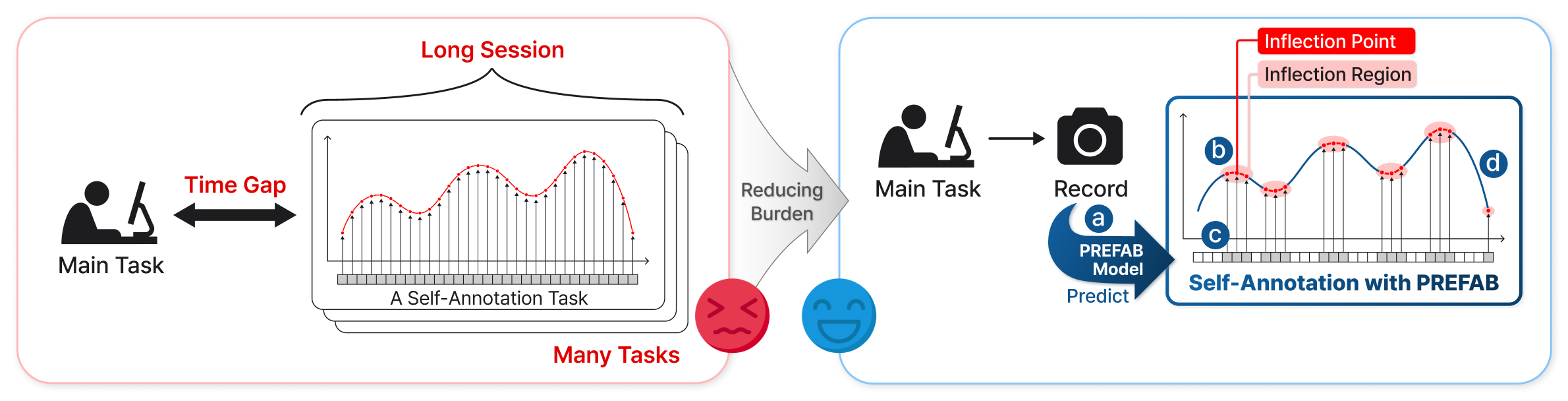}
  \caption{Full self-annotation imposes high cognitive workload (left). PREFAB alleviates this by (a) reconstructing expected affective trends with a preference learning-based model, (b) detecting inflection regions, (c) requiring annotations only for these regions, and (d) interpolating the remainder of the trace (right).}
  \Description{The left panel illustrates traditional self-annotation (full-annotation). Full-annotation increases users’ cognitive load and error rates as the time gap between the main task and annotation grows longer, as the task duration becomes longer, and as the number of tasks increases. The right panel presents the low-budget self-annotation method proposed in this study, PREFAB. After performing the main task, the recorded data is fed into the PREFAB model (a), which reconstructs the affective trajectory and (b) detects inflection regions. (c) Users provide annotations only for these regions, while (d) the remaining parts of the trace are automatically interpolated.}
  \label{fig:teaser}
\end{teaserfigure}
\maketitle

\section{Introduction}
Internal human states---such as frustration, immersion, and arousal---are fundamental notions that affective computing (AC) is tasked to model and predict \cite{affective_computing, affective_computing_review, affective_computing_review_2024}. Such internal states have been widely explored across diverse domains, including art \cite{affective_art}, music \cite{affective_music}, painting \cite{affective_painting}, education \cite{affective_education}, and games \cite{affective_game}. Accurately capturing the temporal dynamics of affective states is of central importance for AC and  the gold standard approach towards that aim is through self-annotation. 
Self-annotation denotes the procedure by which individuals---either during (in-situ) or after performing (retrospective) a task---directly annotate the internal states they experienced in that task \cite{matter_of_anno, if_you_happy}.
While in-situ annotation has the advantage of yielding relatively accurate labels by capturing experiences in real time \cite{matter_of_anno, longterm_fatigue2, belief_and_feeling}, it can interfere with immersion in tasks where concentration is essential---such as games, training, or educational activities---thereby causing side effects that disrupt the primary purpose of the main task \cite{interruption, interruption2}.
In this work, we focus on retrospective self-annotation.
Here, the \textit{main task} refers to active, goal-directed activities such as game-playing, exercise, or training, while \textit{self-annotation} (or first-person annotation) denotes the retrospective self-annotation of one's own affect based on recordings of the main task.

Most prior work on self-annotation in AC focuses on designing tools that capture human affective states as naturally and accurately as possible \cite{RCEA, AffectButton, PAGAN, CARMA, DARMA, RetroSketch, RankTrace, FeelTrace, GTrace, AffectRank}.
For example, tools such as PAGAN \cite{PAGAN} aim to make affect annotation more effective and accessible by providing a web-based platform for crowdsourcing labels for audiovisual content. However, relatively few studies have addressed how to make the annotation process itself more efficient. Inherently, self-annotation relies heavily on human cognitive performance.
The longer the delay between the main task and the self-annotation, the higher the likelihood of memory-related annotation errors \cite{matter_of_anno}. 
In addition, as the duration or the number of annotation tasks increases, the reliability of the annotation task tends to decrease due to user fatigue \cite{longterm_fatigue, longterm_fatigue2}.
Moreover, self-annotation fundamentally requires approximately twice the duration of the main task, as participants must re-experience or review the entire session to provide labels.
To mitigate such cognitive and temporal burdens, several pioneering studies \cite{OCEAN, presup} have explored alternative strategies, most notably by identifying \textit{opportune} intervals based on user's physiological signals and annotating those segments only.
These works produced remarkable results, showing that labels collected only at such opportune moments are not statistically different from those obtained through full scale annotation. Such methods, however, rely heavily on intrusive physiological sensors and do not handle any of the remaining, non-annotated segments.

To address these issues, we introduce \textbf{PREFAB} (\textbf{PREF}erence-based \textbf{A}ffective modeling for low-\textbf{B}udget self-annotation), see Figure~\ref{fig:teaser}. 
PREFAB reduces the cognitive load and time burden of self-annotation using the peak-end rule \cite{peak_end} and ordinal representations of emotion \cite{ordinal_nature}. 
On the one hand, the peak-end rule posits that people’s retrospective evaluations of an experience are disproportionately shaped by its most intense moment (here referred to as the inflection point) and its ending, rather than by the overall average. 
On the other hand, ordinal representations of affective states rely on relative orderings or pairwise preferences rather than on absolute numerical values or categorical labels \cite{ordinal_nature}.

Concretely, PREFAB reconstructs an expected affective trajectory of the main task (see Figure~\ref{fig:teaser}-a) through PREFAB model, detects peaks (i.e. inflection regions) based on this trajectory (see Figure~\ref{fig:teaser}-b), and directs annotators to provide labels only for those regions (see Figure~\ref{fig:teaser}-c) while interpolating the remainder (see Figure~\ref{fig:teaser}-d).
To operationalize the peak-end rule, we segment continuous episodes into event units grounded in the event segmentation theory \cite{EST1, EST2}.
Specifically, the interval between one peak and the subsequent peak is treated as a single affective event, with the moment immediately preceding the next peak---or the end of the episode---regarded as the event’s ``end.''
This segmentation enables consistent mapping of peak-end dynamics within continuous affective trajectories.

To assess the efficiency of the proposed approach, we conducted a technical evaluation of the PREFAB model and devised a user study to address the following four research questions. 

\begin{itemize}
    \item \textbf{RQ1}: Does the ordinal model estimate inflection regions better than other methods, including cardinal modeling?
    \item \textbf{RQ2}: Does our method reduce workload and temporal burden in self-annotation?
    \item \textbf{RQ3}: Do users perceive PREFAB as more effective than full-annotation (conventional approach)?
    \item \textbf{RQ4}: Is annotation quality preserved when annotation is performed only within inflection regions while the remainder of the trace is interpolated?
\end{itemize}

In the technical evaluation performed, we compared the F1 score and time-efficiency alignment ($\Delta$TE) of PREFAB with two na\"ive sampling methods (random and uniform), a heuristic (rule-based event-driven) sampling method, and a cardinal modeling method (regression). 
In the user study, we compared three conditions: Baseline (full-annotation), PREFAB without preview, and PREFAB with preview. 
The preview variant was included to account for individual differences in recognition and comprehension abilities, as short clips alone may not always provide sufficient context for reliable retrospective judgments.
The AGAIN dataset \cite{AGAIN} and the PAGAN annotation tool \cite{PAGAN} are used for this study. In summary, the technical evaluation demonstrates that PREFAB’s ordinal formulation reliably identifies affective inflection regions beyond alternative approaches, and the user study validates that it lowers cognitive load, conditionally mitigates temporal burden, and improves confidence in self-annotation without sacrificing annotation quality.

Our contributions are twofold, spanning both technical and human-centered aspects:

\begin{itemize}
    \item \textbf{Technical Contributions}
    \begin{itemize}
        \item We introduce PREFAB, a novel low-budget self-annotation method that performs preference learning-based selective annotation focusing on peak-end (inflections), and grounded in established theories of ordinal representation and the peak-end rule.
        \item Through model performance evaluation, we demonstrate that the ordinal approach outperforms na\"ive, heuristic, and cardinal modeling methods for modeling affective inflections.
    \end{itemize}

    \item \textbf{Human-Centered (HCI) Contributions}
    \begin{itemize}
        \item We validate PREFAB in a user study of 25 participants, showing that it significantly reduces mental and physical workload, conditionally reduces temporal cost, and increases annotation confidence compared to full annotation, while preserving annotation quality.
        \item We complement these findings with a qualitative interview study, providing in-depth insight into how participants perceived PREFAB and offering design implications for future affective self-annotation systems.
    \end{itemize}
\end{itemize}

\section{Background and Related Work}
\subsection{Self-Annotation in Affective Computing}
\label{subsec: AC}

AC research aims to measure or predict human affective states, and utilize them in applications such as frustration, immersion, arousal, and valence \cite{affective_computing, affective_computing_review, affective_computing_review_2024}. 
A variety of modalities have been explored for affect detection, including physiological signals (e.g., EEG for emotion \cite{affective_eeg}, GSR for stress \cite{affective_gsr}), behavioral signals (e.g., gaze for immersion \cite{affective_gaze}), and in-game behavior \cite{affective_game_survey}. 
Based on these measurement techniques, affective models have been employed across rather diverse domains such as art \cite{affective_art}, music \cite{affective_music}, painting \cite{affective_painting}, education \cite{affective_education}, and games \cite{affective_game,affective_game_survey}. 
For instance, Moon \textit{et al.} \cite{affective_game} proposed a method to adjust the difficulty of a fighting game by predicting players’ affective states. 
They trained player state models to predict four states---challenge, competence, valence, and flow---from game logs, and integrated them into a Monte Carlo Tree Search (MCTS) score function, enabling an AI agent to choose actions that elicit target affective states in players. Therefore, accurately capturing affective states during the main task is a critical prerequisite for affect-driven applications.

To this end, various self-annotation tools have been proposed to effectively capture human affective states \cite{RCEA, ESM, AffectButton, FeelTrace, GTrace, AffectRank, PAGAN, RetroSketch, DARMA, CARMA, RankTrace}. 
For example, FeelTrace \cite{FeelTrace} offers continuous and intuitive arousal–valence labeling by allowing users to move a mouse cursor in real time, instead of selecting discrete state labels.
PAGAN (Platform for Audiovisual General-purpose ANnotation) \cite{PAGAN} provides a web-based infrastructure for crowdsourcing affect labels on audiovisual content. A key feature of PAGAN is its support for relative and unbounded annotation methods (e.g., RankTrace \cite{RankTrace}), which capture perceived changes in affective experience rather than anchoring judgments to absolute scales. 
RCEA (Real-time, Continuous Emotion Annotation) \cite{RCEA} enables participants to provide fine-grained valence–arousal traces while watching mobile videos, offering temporally precise ground truth without increasing cognitive load. 
Most recently, RetroSketch \cite{RetroSketch} introduces a retrospective, trajectory-based approach that allows participants to replay an experience (e.g., VR gameplay) and sketch the temporal course of their emotions, producing continuous annotations that align with ESM and physiological signals. 
These examples highlight the diversity of tools designed to increase the accuracy and naturalness of self-annotation.

Self-annotation can be broadly classified into two categories based on timing: \textit{in-situ} annotation and \textit{retrospective} annotation \cite{if_you_happy, matter_of_anno}. 
In-situ annotation records one’s experience during the main task, whereas retrospective annotation labels the experience after completing the task.
Ecological momentary assessment (EMA) \cite{EMA} and the experience sampling method (ESM) \cite{ESM} are representative examples of in-situ annotation. 
EMA/ESM studies have emphasized the importance of real-time affect capture to mitigate the recall bias of retrospective approaches due to imprecise memory \cite{matter_of_anno, longterm_fatigue2, belief_and_feeling}. 
While some counter studies have shown that interruptions caused by in-situ annotation increase workload, stress, and mental demands \cite{interruption, interruption2}, such approaches are not well suited to active, goal-oriented activities such as games, exercise, training, and education. Consequently, retrospective annotation has been adopted as a more practical alternative in such contexts. In this study, we focus specifically on retrospective annotation.
    
Inherently, retrospective self-annotation relies heavily on human cognitive performance: the longer the delay between the main task and annotation, the higher the likelihood of memory-related errors \cite{matter_of_anno}, and prolonged or repeated annotation tasks tend to reduce reliability due to user fatigue \cite{longterm_fatigue, longterm_fatigue2}. 
To mitigate these issues, some studies have explored detecting ``opportune'' intervals from physiological signals and limiting annotation to those segments. 
Akhilesh \textit{et al.} \cite{OCEAN} collected participants’ biosignals, applied $k$-means clustering to predict opportune moments, and validated them against continuous annotations in the CASE dataset \cite{CASE}, showing that selective annotations were statistically comparable. 
Building on this, Swarnali \textit{et al.} \cite{presup} trained an LSTM-based model \cite{LSTM} on physiological signals from participants and had the same participants perform both full-sequence and selective annotation of the predicted intervals in the CASE dataset, again demonstrating statistical consistency. These studies suggest that accurate annotations for key moments can be obtained without full-sequence labeling. 
However, both approaches share several limitations as they do not consider the non-opportune intervals, they rely on the collection of external physiological signals, and they model affect in a cardinal way.

\subsection{Theoretical Foundations}

To reduce annotator burden and improve human affect modeling accuracy in self-annotation, this study leverages two established theoretical frameworks that are detailed here: the peak-end Rule (see Section~\ref{subsec:peakendrule}) and the ordinal representation of emotions (see Section~\ref{subsec:ordinalnature}).

\subsubsection{Peak-End Rule}
\label{subsec:peakendrule}
The peak-end rule \cite{peak_end} is a psychological principle derived from economics, which posits that people's retrospective evaluations of an experience are largely determined by the emotional intensity at its peak (the most extreme point) and at its end, rather than by the overall average. 
This heuristic has been consistently observed across various domains, including multimedia preference \cite{preference_peakend}, daily life logging \cite{if_you_happy}, and education \cite{education_peakend}. 
In particular, Scharbert \textit{et al.} \cite{if_you_happy} showed that peak-end bias strongly influences retrospective annotations.
While they did not directly advocate selective labeling, their findings imply that reliable affect labels may be obtained by annotating only peak and end moments (inflection regions), which in turn motivates our approach.

Conventional studies of the peak-end rule typically focus on single, discrete experiences, whereas our study deals with continuous episodes composed of multiple, overlapping events. In such contexts, defining clear ``peaks'' and ``ends'' becomes non-trivial.
To operationalize this, we adopt a na\"ive assumption inspired by the event segmentation theory \cite{EST1, EST2}, which describes how the human brain parses continuous experience into meaningful events based on prediction errors and contextual shifts.
Following this view, we treat each inflection point---where affective intensity changes abruptly---as both the peak of a new event and the end of the preceding one, such that the interval between consecutive peaks constitutes a single affective event.

A potential concern, however, is whether annotating only such inflection regions, independently of their surrounding context, can yield results comparable to full-sequence annotation. Evidence from related work partially addresses this concern. 
As described in Section~\ref{subsec: AC}, prior studies \cite{OCEAN, presup} have predicted ``opportune'' intervals from physiological signals and found that selective annotations at those intervals were statistically consistent with full annotations. While these works were not based on peak-end theory, their findings support the feasibility of selective annotation by showing that accurate labels can be obtained from sparse key moments.

\subsubsection{Ordinal Nature and Preference Learning}
\label{subsec:ordinalnature}

Yannakakis \textit{et al.} \cite{ordinal_nature} emphasized that human affective judgments are inherently subjective and relative (ordinal) rather than absolute (cardinal). This perspective suggests that it is more effective to model human affective experience through relative comparisons between task segments (e.g., whether one moment is more arousing than another), rather than by predicting absolute affective values for individual moments. Preference learning (PL) embodies this learning task by training models to capture the comparative relationship between pairs of segments, and has been widely applied in affect modeling \cite{preference_peakend, PreferenceLearning_Yanna, PreferenceLearning_Yanna2, RankNEAT, Pixels_and_Sounds, AGAIN}. 
A representative implementation of PL is RankNet \cite{burges2005ranknet}, which formulates preference prediction as a binary classification problem: given two input segments $x_i$ and $x_j$, the model ($\Phi$) learns to predict whether $x_j$ is greater than $x_i$ as follows:

\begin{equation}
\label{eq:logistic}
\begin{aligned}
    \Phi(x_i) & = p_i,~\Phi(x_j) = p_j,~p_j - p_i = p_{ij}, \\
    P(y_{ij} = 1) & = P(p_j \succ p_i) = \sigma(p_{ij}) = \frac{1}{1+e^{-p_{ij}}}.
\end{aligned}
\end{equation}

\begin{equation}
\label{eq: ranknet loss}
    \text{BCE}(x_i, x_j) = -y_{ij} \log(\sigma(p_{ij})) - (1 - y_{ij}) \log(1 - \sigma(p_{ij})).
\end{equation}

\noindent where $y_{ij}=1$ indicates that segment $x_j$ is preferred over segment $x_i$, and the network is trained with a binary cross-entropy (BCE) loss to align predictions with true pairwise preferences. 

However, binary preference learning assumes a strict ordering and cannot naturally represent cases where two segments are judged as equal, which is common in affect annotation. 
To address this limitation, we extend the framework to a three-class setting: greater ($class=2$), equal ($class=1$), and less ($class=0$). For this purpose, we adopt the ordinal cross-entropy (OCE) loss function \cite{pedregosa2017OCE}, which models class probabilities using cutoff points $c_0$ and $c_1$:

{\small
\begin{equation}
\label{eq:OCE}
    P(y_{ij} = \text{class}) = 
    \begin{cases}
        \sigma(c_0 - p_{ij}), & \text{class}=0,~p_i \succ p_j \\
        \sigma(c_1 - p_{ij}) - \sigma(c_0 - p_{ij}), & \text{class}=1,~p_i = p_j \\ 
        1 - \sigma(c_1 - p_{ij}), & \text{class}=2,~p_i \prec p_j.
    \end{cases}
\end{equation}
}

Based on this probability formulation, the OCE loss is defined as the negative log-likelihood over the true class label:

\begin{equation}
\label{eq:OCE_loss}
{\text{OCE}}(x_i, x_j)
= -\log P(y_{ij}\mid p_{ij}, c_0, c_1).
\end{equation}

\begin{figure}[!tb]
  \includegraphics[width=0.95\columnwidth]{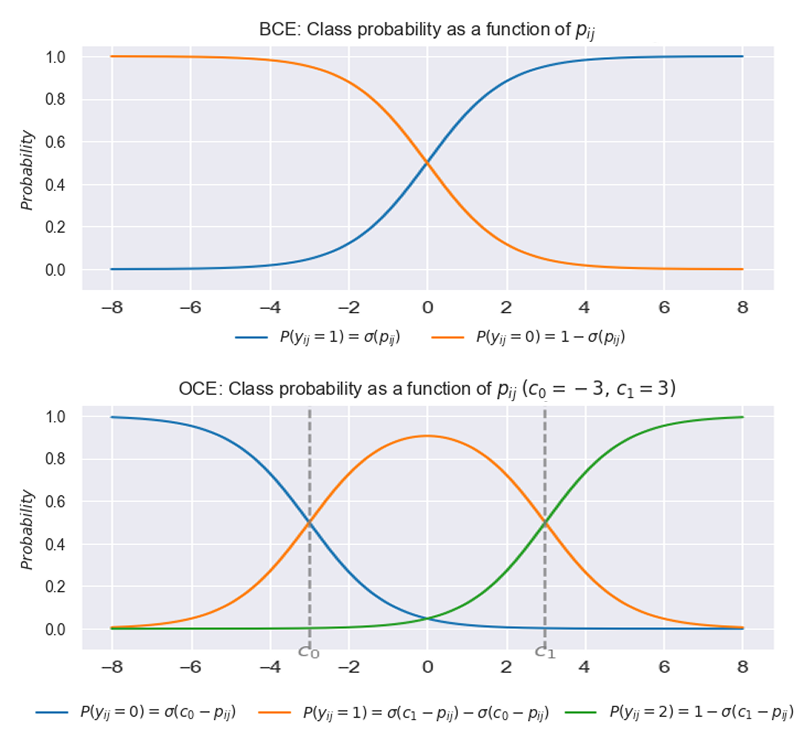}
  \caption{Visualization of BCE and OCE probability functions. BCE produces a single sigmoid transition between two classes, while OCE partitions the latent score $p_{ij}$ into three ordinal regions using cutpoints $c_0$ and $c_1$.}
  \Description{Two side-by-side line charts comparing BCE and OCE probability curves.
The BCE plot shows two sigmoid-shaped curves crossing at the midpoint:
a blue curve for $P(y_{ij}=1)=\sigma(p_{ij})$ rising from 0 to 1, and an orange curve for $P(y_{ij}=0)=1-\sigma(p_{ij})$ falling from 1 to 0.
The OCE plot shows three smooth curves for OCE: a blue curve decreasing for $P(y_{ij}=0)=\sigma(c_0-p_{ij})$, an orange curve peaking in the center for $P(y_{ij}=1)=\sigma(c_1-p_{ij})-\sigma(c_0-p_{ij})$, and a green curve increasing for $P(y_{ij}=2)=1-\sigma(c_1-p_{ij})$.
Vertical dashed lines mark the cutpoints $c_0=-3$ and $c_1=3$, labeled below the axis.
Together, the plots illustrate how OCE generalizes BCE by introducing an intermediate ``equal'' region between the two extremes.
}
  \label{fig:BCEOCE}
\end{figure}

This formulation generalizes the binary logistic model into an ordinal cumulative link model, in which the logit difference $p_{ij}$ is smoothly partitioned by ordered cutpoints $c_0 < c_1$. These cutpoints are predefined constants determined by the experimenter.
Figure~\ref{fig:BCEOCE} illustrates how the probability differs between BCE and OCE.
Each class probability corresponds to the likelihood that $p_{ij}$ falls within a specific interval on this latent scale, yielding a continuous and differentiable representation of ordinal relationships.
Conceptually, OCE preserves the comparative structure of preference learning while introducing a neutral (equal) class between the two extremes, thereby offering a more faithful and psychologically interpretable model of human affective judgment.

\section{Study 1: Technical Approach}
This section introduces the core technical aspects of PREFAB, consisting of four steps:
\begin{itemize}
    \item \textbf{Step 1 (1-1 and 1-2): Prediction of Affective Change (see Figure~\ref{fig:teaser}-a)}: The recorded user data from a main task is fed into the trained PREFAB model to predict a graph representing the change in the target affect.
    \item \textbf{Step 2: Inflection Point Detection (see Figure~\ref{fig:teaser}-b)}: PREFAB identifies inflection points in the reconstructed affect change curve, and defines inflection regions using a given window size.
    \item \textbf{Step 3: Selective Annotation (see Figure~\ref{fig:teaser}-c)}: The user performs self-annotation only for the identified inflection regions.
    \item \textbf{Step 4: Interpolation see (Figure~\ref{fig:teaser}-d)}: The system interpolates the remaining unannotated regions to complete the affective trajectory.
\end{itemize}

\subsection{Step 1-1: Data Preprocessing}
\subsubsection{Dataset}
In this study, we used the AGAIN dataset \cite{AGAIN} which contains play log data, gameplay videos, self-annotated arousal data, and biographical information from over 120 participants who played nine different games. 
The game log data was cleaned to provide a frame rate of 4 timesteps per second. 
The self-annotated arousal data was collected using the PAGAN tool \cite{PAGAN}, which records relative changes in arousal during each game. The dataset provides both general game features and unique game-specific features, both of which were utilized in our model as proposed by the original dataset authors.

\subsubsection{Biographical Data Preprocessing}
To account for individual subjective characteristics, we quantized the biographical data from the AGIAN dataset and used it as conditional input to the PREFAB model.
This data includes a participant's age, gender, nationality, dominant hand, gaming frequency, gamer type, preferred game platform, and favorite games.
Specifically, there were 86 different favorite game titles in the original biography data.
We converted them into 32 game genres using game genre codes provided by MetaCritic services \cite{metacritic}.
This was done because quantizing individual game titles as separate labels causes games that share similar characteristics to be treated as different categories, thereby reducing the representational fidelity of gaming preferences (e.g., \textit{American Truck Simulator vs. Euro Truck Simulator, and Doom 2 vs. Doom Eternal}).

\subsubsection{Auxiliary Task}
\label{subsec: auxtask}

To facilitate the learning of a more robust representation for arousal changes, we introduced an auxiliary classification task. 
Auxiliary objectives are widely employed in deep learning to aid representation learning, particularly when primary supervision is localized, biased, or unstable \cite{aux1, aux2, aux3}. By enforcing relevant latent structures, these objectives facilitate more stable learning and improved generalization.

In our context, relying solely on local comparisons risks ignoring broader temporal dynamics. For instance, an identical combat scene can elicit different arousal responses depending on when it occurs: a player might be relaxed in the early game but highly tense under time pressure in the late game. Furthermore, players often become desensitized as the session progresses. To capture these global dependencies, we introduce an auxiliary classification task designed to recognize coarse-grained arousal patterns, complementing the local supervision of the main objective.

\begin{figure}[!tb]
  \includegraphics[width=\columnwidth]{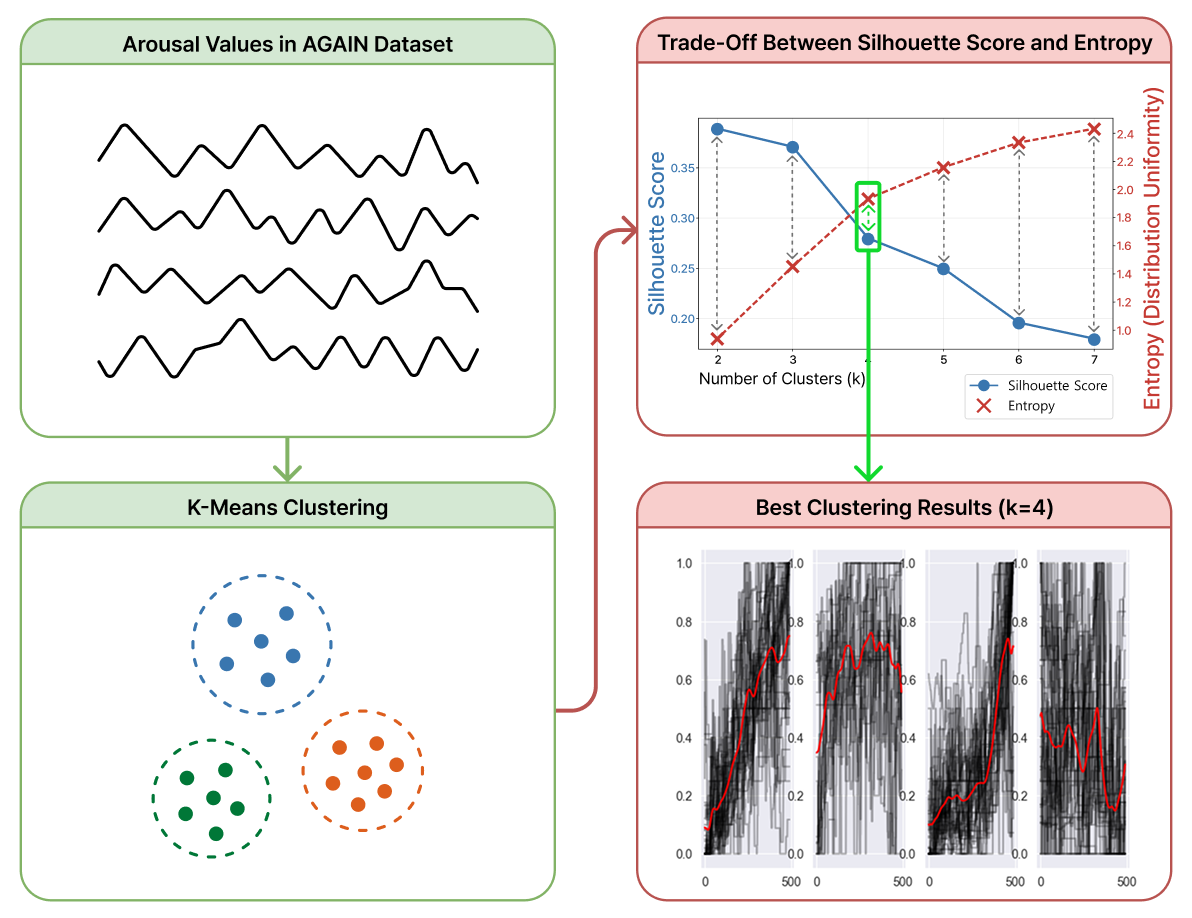}
  \caption{Evaluation and results of dynamic time warping (DTW)-based clustering. The left part of the figure shows the trade-off between the cluster silhouette score and data distribution (entropy). The right part of the figure visualizes the four resulting clusters of arousal change patterns ($k=4$): individual time series and mean values are depicted in  in gray and red color, respectively.}
  \Description{This figure illustrates the process of generating labels for the auxiliary task. Arousal graphs from the AGAIN dataset (top left) are clustered using k-means based on dynamic time warping distance (bottom left). The number of clusters, k, is chosen from 2 to 7 by balancing entropy, which reflects the uniformity of data distribution, and the silhouette score, which reflects the separability of k-means clustering. The optimal value is determined as $k=4$ (top right). The bottom right panel shows the final clustering results.}
  \label{fig:dtw}
\end{figure}

Figure~\ref{fig:dtw} shows how we formulated the auxiliary task through Dynamic Time Warping  (DTW) \cite{DTW} clustering. We clustered the time-series arousal data from the AGAIN dataset using DTW distance as the metric for k-means clustering. Then, we evaluated the optimal number of clusters ($k=2-7$) by assessing the balance between the data sample distribution across clusters by entropy and the silhouette score, which measures how well-separated the clusters are. 
Finally, we found that $k=4$ achieved the best balance between cluster separation (high silhouette score) and distribution uniformity (high entropy), yielding four distinct clusters of sufficient size without being overly fragmented or imbalanced.
These clustering results were subsequently used as labels ($c$) for the auxiliary task, which classifies each game segment according to its overall arousal-change pattern.

\begin{figure*}[!tb]
  \includegraphics[width=\textwidth]{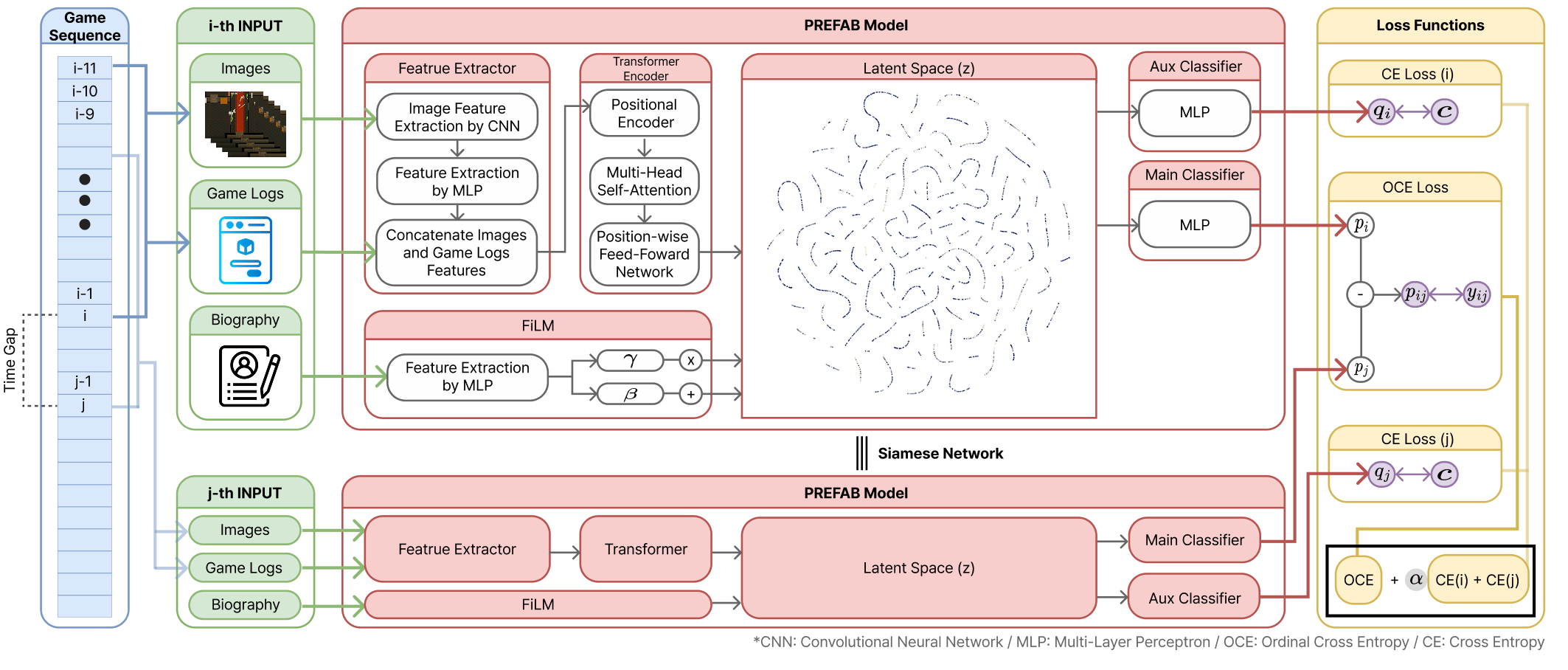}
  \caption{Architecture of the PREFAB model. The model takes two consecutive segments sampled at a 1-second interval, each consisting of 12 image frames, game log features, and the player’s biographical information. Both segments are processed by the same encoder in a Siamese network structure. In the inference stage, a single segment is fed into the model to predict the relative value $p$ at each time point.}
  \Description{Diagram of the PREFAB model architecture. On the left, consecutive game sequences are arranged with time intervals, and each sequence consists of image frames, game logs, and biography information. These inputs are fed into the feature extraction modules. Images are compressed into vectors through CNNs and MLPs, while game logs are compressed through MLPs. The vectors are then concatenated and passed into a Transformer encoder (positional encoding, multi-head self-attention, and position-wise feedforward network). Biography information is incorporated as conditional information through a FiLM module, which rearranges the latent space ($\gamma$, $\beta$) to reflect individual characteristics. The resulting latent space (z) is connected to both the main classifier and the auxiliary classifier. The main classifier predicts relative arousal values, while the auxiliary classifier predicts trend clusters for the auxiliary task. The two input sequences (i, j) are processed by identical encoders, and the difference between their outputs is used to compute the Ordinal Cross Entropy (OCE) loss. Each classifier also applies a Cross Entropy loss, and the final loss function is defined as a combination of the OCE and the two CE losses.}
  \label{fig:prefab module}
\end{figure*}

\subsubsection{Paired Data Formulation (leftmost blue box in Figure~\ref{fig:prefab module})}
Following a common approach in prior research \cite{AGAIN, camilleri2017towards, RankTrace, 3sec1, 3sec2, 3sec3, 3sec4}, we used a 3-second window with a 1-second temporal gap to construct data pairs for preference learning.
Each input segment $X_i$ consists of a user's biographical data ($b$), 12 frames (3 seconds) of game images ($I_{i-11}, \dots, I_i$), and corresponding game log features ($F_{i-11}, \dots, F_i$) at time index $i$. This can be expressed as:
\begin{equation}
   X_i = \big(\{F_{i-11},\dots,~F_i\},~\{I_{i-11},\dots,~I_i\},~b \big),\quad i>12 
\end{equation}

Arousal at time $i$ is denoted as $A_i$.
For preference learning, we define the paired label $Y_{ij}$ as a three-class variable based on the temporal change of arousal between $i$ and $j=i+4$ (i.e., one second later):
\begin{equation}
Y_{ij} =
\begin{cases}
1 & \text{if } A_j - A_i > 0 \quad \text{(increase)} \\
0 & \text{if } A_j - A_i = 0 \quad \text{(no change)} \\
-1 & \text{if } A_j - A_i < 0 \quad \text{(decrease)}
\end{cases}
\end{equation}

\subsection{Step 1-2: PREFAB Model Implementation}

\subsubsection{Model Architecture}
Figure~\ref{fig:prefab module} demonstrates the overall PREFAB model architecture.
The model takes two sequential data pairs as input: game logs, screen captures, and user biographical data. 
It is built on a transformer backbone \cite{vaswani2017attention} and uses a \textit{Feature-wise Linear Modulation} (FiLM) method \cite{FiLM} to condition its latent space on individual user characteristics, allowing for the prediction of personalized arousal patterns. Additionally, we integrated an auxiliary task to help the model learn more distinct data representations (See Section~\ref{subsec: auxtask}). 
The effects of the FiLM method and the auxiliary task used in our model are detailed in Appendix~\ref{adx:ablation}.

The PREFAB model is implemented as a Siamese network structure \cite{siamese} to process two sequential input segments, $i$-th and $j$-th, simultaneously (see Figure~\ref{fig:prefab module}).
First, the raw image and game feature data are fed into separate feature extractors. 
A convolutional neural network (CNN) and several multi-layer perceptrons (MLP) compress these data into low-dimensional vectors.
These vectors are then concatenated and passed to the next stage. Second, we adopted the encoder component of a transformer \cite{vaswani2017attention}, which is well suited for modeling time-series data. 
The encoder was configured with an input dimension of 200 and composed of two stacked layers to process the concatenated vectors.
Third, to reflect personal information into the model, a FiLM method is applied to the transformer's output.
This method rearranges the model's latent space ($z$), allowing it to capture unique arousal patterns for each individual.
Finally, the model concludes with two separate MLP heads: the main classifier and the auxiliary classifier.
The main classifier predicts the relative affect value of the input segment, while the auxiliary classifier predicts the affect trend cluster as defined in Section~\ref{subsec: auxtask}.
During training, the relative values derived from the Siamese network are subtracted to obtain the affective change trend.

At inference time, however, the PREFAB model is applied to single segments rather than pairs. 
Instead of explicitly computing differences between two segments, the model outputs the relative arousal value for each segment individually. 
By iterating this process across the entire sequence, the predicted relative values are aggregated to reconstruct the full arousal-change trajectory (see Figure~\ref{fig:teaser}-a).

\subsubsection{Loss Functions}
For the main downstream task, each segment is passed through the main classifier to produce a scalar representing its relative arousal value ($p_i$ and $p_j$). During training, the difference between two segments, $p_{ij} = p_j - p_i$, is used together with the ground-truth label $y_{ij}$ (i.e. increase / no change / decrease) to compute the ordinal cross-entropy (OCE) loss \cite{pedregosa2017OCE}. 
The cut-points for the OCE were set to [-1, 1]. 
For the auxiliary task, each segment is classified into one of four arousal trend clusters ($c$) using a standard cross-entropy (CE) loss. Here, $q_i$ and $q_j$ denote the predicted probability distributions over the four clusters for segments $i$ and $j$. 
To prevent the auxiliary task from dominating optimization, we applied a scaling factor ($\alpha = 0.001$) to its loss term. The overall training objective is defined as:
\begin{equation}
    \label{eq:total loss}
    L = \text{OCE}(y_{ij}, p_{ij}) + \alpha \cdot \big( \text{CE}(c, q_i) + \text{CE}(c, q_j) \big).
\end{equation}

\subsection{Step 2: Inflection Detection}
\label{subsec:inflection_detection}
We used the \textit{find\_peaks} function of scipy library \cite{scipy} to identify the inflection points from the arousal graph reconstructed by the PREFAB model (see red dots in Figure~\ref{fig:teaser}-b). All hyperparameters were set at their default values. The inflection points include not only local maxima, but also local minima. Since the \textit{find\_peaks} function is designed to detect local maxima, we applied it to the inverted arousal graph to also capture local minima, which were then included as inflection points.

A potential issue with this approach is that the \textit{find\_peaks} function may miss subtle changes in the ground truth graph, especially in long flat regions or when the slope is too gentle. To address this limitation, we developed a complementary rule: any segments with a significant change in gradient that were not identified as inflection points by \textit{find\_peaks} were also included.

Each identified inflection point became the center of a 5-second annotation clip (i.e. the inflection region), extending 2.5 seconds before and after the point (see red area in Figure~\ref{fig:teaser}). 
We defined a 5-second window to balance annotation reliability and cognitive constraints. Prior work has shown that retrospective self-annotation typically involves a reaction lag of about 1–3 seconds between perception and labeling \cite{time_lag1, time_lag2, time_lag3}, and humans generally require at least 3 seconds of exposure to comprehend the main content of a video segment \cite{clip_duration}. 
Thus, we assume that a 5-second window provides a ``lower bound'' that is both cognitively feasible and sufficient for recognizing contextual content, while still short enough to avoid excessive workload. If the inflection region from adjacent inflection points overlaps, we merge the two to form a single, longer region. Finally, in \textbf{[Step 3]} users were asked to perform self-annotation only for these final selected clips (see Figure~\ref{fig:teaser}-c).

\subsection{Step 4: Interpolation}
\label{subsec:interpolation}

\begin{algorithm}[!tb]
\caption{Interpolation Method for Unannotated Regions via Cumulative Slope Propagation}
\label{alg:interpolation}
\begin{flushleft}
\textbf{Input:}
\begin{itemize}
\item Total sequence length $T$ with time indices $t = 1,\dots,T$.
\item Inflection regions $\{R_k = [s_k, e_k]\}_{k=1}^n$ in temporal order ($e_k < s_{k+1}$).
\item Annotated arousal values $A_{\tau}$ available only for $\tau \in R_k$ (for each $k$).
\end{itemize}
\textbf{Output:} Interpolated arousal sequence $\hat{A}_t$ for $t = 1,\dots,T$.
\end{flushleft}
\vspace{1mm}
\begin{enumerate}[leftmargin=1.5em]
  \item \textbf{Initialize.} Set $\hat{A}_t \gets 0$ for all $t=1,\dots,T$.
  \item \textbf{For each annotated region $R_k = [s_k, e_k]$:}
  \begin{enumerate}
    \item (\emph{Apply offset and add annotated values})
          Let $\text{offset}_k = \hat{A}_{s_k}$.
          For all $\tau \in [s_k, e_k]$, update
          \[
            \hat{A}_{\tau} \gets \text{offset}_k + A_{\tau}.
          \]
    \item (\emph{Compute average slope of latter half})
          Let $m_k = \lfloor (s_k + e_k)/2 \rfloor$, then
          \[
            \text{slope}_k = \frac{1}{\,e_k - m_k\,}
            \sum_{\tau = m_k}^{e_k - 1} \big(A_{\tau+1} - A_{\tau}\big).
          \]
    \item (\emph{Define propagation index set})
          \[
            \tilde{\mathcal T}_k =
            \begin{cases}
              \{e_k{+}1, \dots, s_{k+1}\}, & k < n,\\
              \{e_k{+}1, \dots, T\}, & k = n.
            \end{cases}
          \]
    \item (\emph{Cumulative propagation})
          Set the starting anchor $\hat{A}_{e_k}$ (already set in step 2a).
          For each $\tilde{t} \in \tilde{\mathcal T}_k$ in ascending order:
          \[
            \hat{A}_{\tilde{t}} \gets \hat{A}_{\tilde{t}-1} + \text{slope}_k.
          \]
  \end{enumerate}
\end{enumerate}
\end{algorithm}

The PREFAB method identifies and annotates only the inflection regions, leaving the rest of the sequence unannotated. 
Motivated by the peak-end rule, we assume that affective experience in non-annotated periods can be reasonably approximated by interpolating between surrounding inflection regions. 
For simplicity purposes we adopt linear interpolation to estimate the values for these unannotated segments (see Figure~\ref{fig:teaser}-d).
Algorithm~\ref{alg:interpolation} demonstrates the procedure. Specifically, we first calculate the slope of the final half of an annotated clip. This slope is then used to linearly propagate the arousal change until the beginning of the next annotated clip. 
This propagated value serves as an offset for the annotated values of the subsequent inflection region. This process is repeated for each unannotated segment until the end of the recorded task.

\section{Result 1: Model Performance Evaluation (RQ 1)}
\label{sec: technical evaluation}
We first examined whether PREFAB model more effectively predicts inflection regions compared to na\"ive, heuristic, and cardinal model-based approaches.
To ground the comparison, we used self-annotated arousal curves from the AGAIN dataset as ground truth (GT). 
Across all nine games, PREFAB provided the best combination of F1 score and time-efficiency alignment ($\Delta$TE).
These results support RQ1, showing that an ordinal preference-learning approach can predict inflection regions more effectively than na\"ive, heuristic, or cardinal alternatives.

\subsection{Metrics}
Evaluation focused on two metrics: region-level F1 and $\Delta$TE. 
The first metric is region-level F1, which captures how well the predicted inflection regions overlap with those of the GT.
The F1 score is calculated based on the overlap between the predicted inflection regions and the inflection regions calculated from the ground truth directly.

\begin{equation}
    F1 = 
    \frac{2 (\,\mathcal{I}_{\mathrm{GT}} \cap \hat{\mathcal{I}}\,)}
         {\mathcal{I}_{\mathrm{GT}} + \hat{\mathcal{I}}}
\end{equation}

\noindent where $\mathcal{I}_{\mathrm{GT}}$ is the inflection regions calculated by applying the inflection detection algorithm described in Section~\ref{subsec:inflection_detection} to the GT arousal curves; and $\hat{\mathcal{I}}$ are the predicted inflection regions.

However, F1 alone can be misleading in two cases:
(i) \textbf{over-coverage}, when the predicted regions cover almost the entire timeline, causing many overlaps and producing high recall at the cost of precision; and 
(ii) \textbf{under-coverage}, when the predicted regions are very few and narrow but happen to fall inside the regions of $\mathcal{I}_{\mathrm{GT}}$, which can also yield high precision at the cost of recall. 
In both cases, a method may achieve a deceptively high F1 score without actually reproducing GT’s selective annotation behavior.

To address these issues, we use $\Delta$TE, which measures how closely a method's time efficiency ($\hat{\mathrm{TE}}$) matches the GT’s time efficiency (TE$_{GT}$) of inflection-region sampling.
\begin{equation}
\label{eq:deltaTE}
    \Delta \text{TE}
    = \frac{1}{N} \sum_{i=1}^{N}
      \left|\,\text{TE}_{GT,i} - \hat{\mathrm{TE}}_{i}\,\right|.
\end{equation}

TE is computed as:
\begin{equation}
\label{eq:te}
    \text{TE} = 
    \frac{\text{Total duration} - \sum \mathcal{I}}
         {\text{Total duration}}.
\end{equation}

\subsection{Baselines}
We compared PREFAB against three categories of baselines, chosen to reflect both trivial and representative alternatives.
First, we included two na\"ive baselines—\textbf{uniform} and \textbf{random} sampling—that select points without any task knowledge, while matching the average number of GT inflections per game.
These provide chance-level performance boundaries against which more informed methods can be meaningfully evaluated.
Second, we implemented a \textbf{rule-based} heuristic baseline that samples at score rise and fall moments derived from game logs. This baseline does not rely on large-scale model training; instead, it represents a lightweight, knowledge-driven alternative that can be readily implemented when domain expertise is available.
Finally, we included a cardinal \textbf{regression model} that predicts absolute arousal trajectories to estimate inflection points. 
To ensure a fair comparison, the regression model was implemented with the same architecture and hyperparameters as PREFAB, differing only in its training objective: cardinal prediction of absolute arousal values rather than ordinal preference learning. 
As cardinal modeling is widely used in affective computing \cite{affective_computing, affective_computing_review, affective_computing_review_2024}, this baseline provides the primary model-based comparison point.

\subsection{Quantitative Results and Visual Evidence}
\begin{figure}[ht]
  \includegraphics[width=0.95\columnwidth]{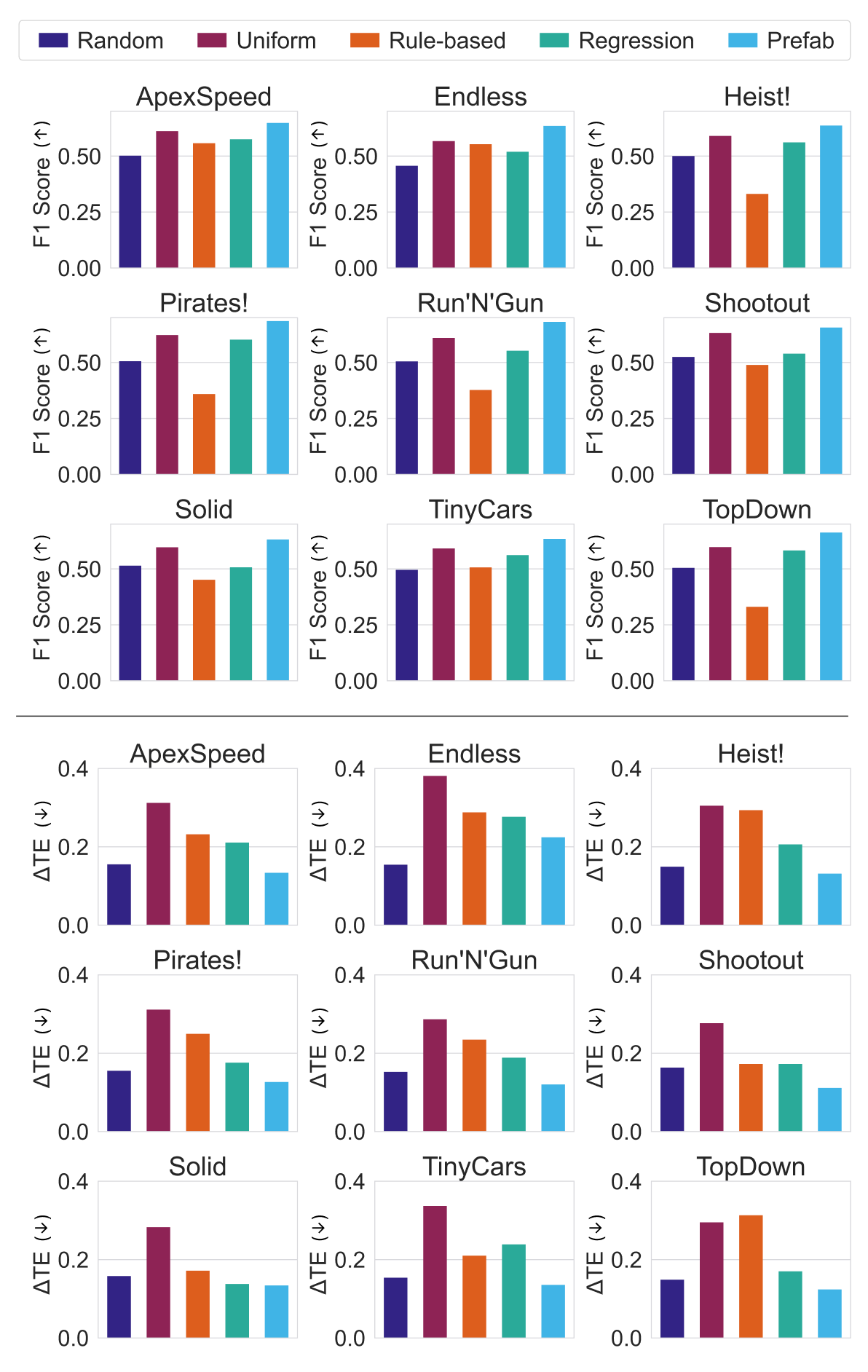}
  \caption{Model performance across nine games. The top panels (3×3) report F1 scores (higher values are better) for each method, and the bottom panels (3×3) report corresponding $\Delta$TE (lower values are better).}
  \Description{Figure compares model performance across nine games (ApexSpeed, Endless, Heist!, Pirates!, Run’N’Gun, Shootout, Solid, TinyCars, TopDown). The top set of 3×3 bar charts shows F1 scores for five methods (Random, Uniform, Rule-based, Regression, PREFAB). PREFAB consistently achieves the highest F1 in all games. The bottom set of 3×3 bar charts shows $\Delta$TE for the same methods. PREFAB’s $\Delta$TE values are consistently closest to the GT baseline compared to other methods with the exception for Endless game (second closest).}
  \label{fig:performance}
\end{figure}

Figure~\ref{fig:performance} summarizes F1 and $\Delta$TE across all games. 
On \textbf{F1}, PREFAB consistently outperformed all baselines with statistically significant differences. 
An ANOVA confirmed significant differences among methods for both metrics ($p < .001$; see details in Table~\ref{tab:performance}), and Bonferroni-corrected pairwise t-tests indicated that all method pairs differed significantly (all $p < .001$).

On \textbf{$\Delta$TE}, PREFAB again showed the closest approximation to GT on average, with the exception of the Endless game, where it ranked second.
On average, the $\Delta$TE$_{PREFAB}$ (0.138 $\pm$ 0.033) was smaller than all other methods--random (0.154 $\pm$ 0.004), regression (0.197 $\pm$ 0.041), rule-based (0.241 $\pm$ 0.051), and uniform (0.310 $\pm$ 0.032). 

\begin{figure*}[t]
  \includegraphics[width=0.8\textwidth]{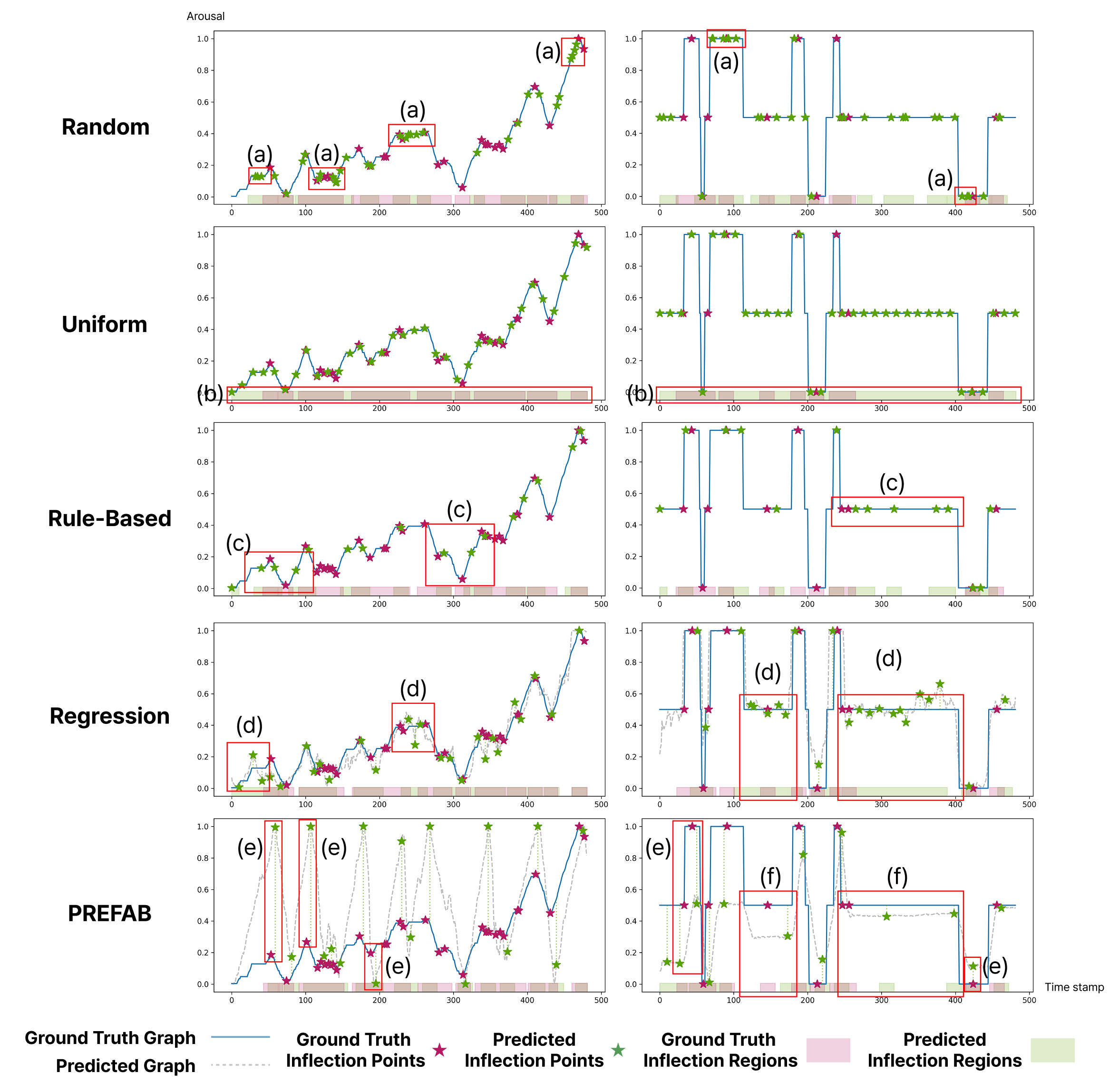}
  \caption{Comparison of inflection point sampling across methods. X-axis indicates timestamps and Y-axis arousal levels. Green stars mark inflection points predicted by each method, with green shaded regions denoting their expanded inflection regions. Red stars and regions indicate the corresponding ground truth (GT). For model-based methods (PREFAB and regression), dashed lines show reconstructed trajectories from trained models.}
  \Description{Figure compares inflection point sampling across five methods. Random sampling matched the number of ground-truth (GT) points but often placed them in clustered positions, leaving large gaps. Uniform sampling produced artificially high F1 by covering nearly the entire timeline, resulting in near-zero time efficiency. The rule-based method followed score changes, but misalignment with affective dynamics led to frequent misses of GT regions. Regression, trained on absolute arousal values, generated noisy trajectories with excessive inflection points, especially in flat segments. In contrast, PREFAB, trained on ordinal relationships, produced trajectories that more faithfully aligned with true GT inflections, yielding higher region overlap and stable performance.}
  \label{fig:reconstruction}
\end{figure*}

Figure~\ref{fig:reconstruction} offers a closer look at how regions were captured, so that why these results were derived.
The left panel shows a game where the number of GT inflection points was close to average, and the right a game with sparse GT inflection points.
Each row represents the sampling results performed by each baseline and PREFAB.
For the model-based methods methods (PREFAB and Regression), we first predicted the expected arousal trajectory using the trained model, and then detected inflection points from the reconstructed graph.
Accordingly, the reconstructed graph is also shown as a gray dashed line on the Figure~\ref{fig:reconstruction}.
In contrast, the Rule-based, Uniform, and Random methods sampled inflection points directly from game logs without reconstructing a graph.
Thus, no predicted graph is displayed.

For random sampling, the number of inflection points was matched to GT averages, but their positions were chosen blindly. 
As shown in Figure~\ref{fig:reconstruction}.a, the sampled points often clustered together, leaving other regions empty. 
This explains why the method produced F1 scores around or below 0.5 across games, indicating poor region overlap, even though its $\Delta$TE ($0.154\pm0.043$) was the second-lowest overall. 
In other words, its apparent closeness to GT efficiency is misleading, because the low deviation results mainly from clustered samples rather than meaningful alignment with GT, causing poor F1 score.

For uniform sampling, once each sampled point is expanded into a region, it ends up covering almost the entire timeline (Figure~\ref{fig:reconstruction}.b). 
This naturally produces a high F1 score, since every GT region is included somewhere in the coverage. 
However, time efficiency dropped to nearly zero, meaning that annotators would still need to label almost the entire sequence.
This explains why uniform achieved the second-highest F1 across games; yet this score reflects trivial coverage of the entire timeline rather than selective accuracy, which is confirmed by its worst $\Delta$TE ($0.310\pm0.032$).

The rule-based baseline provides a more plausible strategy by looking at score rises and falls. 
This helped it avoid covering the entire timeline, and in some cases it captured broad trends when score changes loosely aligned with arousal changes. 
However, because in-game events do not perfectly match affective inflections, the detected regions often missed GT (Figure~\ref{fig:reconstruction}.c), resulting in consistently lower F1 than model-based methods and a relatively large deviation from GT TE ($\Delta$TE $=0.241\pm0.051$).

The contrast between cardinal and ordinal modeling methods is particularly revealing. 
Because the regression model is trained to predict ``absolute'' values, its decision surface must sensitively fit any variation in the input. 
As a result, even small changes in the input can induce spurious fluctuations in the predicted trajectory, which inflate the number of detected inflection points (Figure~\ref{fig:reconstruction}.d).
PREFAB, by contrast, does not aim to reproduce absolute curve values; instead, it focuses on ordinal preferences between segments. 
While its reconstructed curve may appear less similar to the global GT shape, it aligns more faithfully with the true inflection moments, yielding much higher region overlap (PREFAB predictions in Figure~\ref{fig:reconstruction}.e). 
This advantage is especially evident in long flat segments, where regression over-samples due to noise sensitivity, but PREFAB remains stable (Figure~\ref{fig:reconstruction}.d and f).

\section{Study 2: User Study for Practical Evaluation}

\begin{table}[ht]
\begin{center}
  \caption{Research Design.}
  \Description{Table summarizes the research design, organized into independent, dependent, and control variables. The independent variable is the annotation method, with three conditions: baseline full-annotation, PREFAB without preview, and PREFAB with preview. Dependent variables include quantitative evaluation (self-reported workload and confidence, and self-reported annotation quality) and qualitative evaluation (post-interview). Control variables include the game (TopDown from the AGAIN dataset), the tool (PAGAN), participants (within-subject design with 25 participants), interpolation policy (linear interpolation), and experimental order (Latin-square counterbalanced order for both annotation and evaluation sessions).}
  \label{tab:exp var}
  \resizebox{\columnwidth}{!}{%
  \begin{tabular}{c|c|c}
    \toprule
    Variables & \multicolumn{2}{c}{Components} \\
    \midrule
    \multirow{3}{*}{Independent} & \multirow{3}{*}{Annotation Method} & Baseline (Full-Annotation) \\
    \hhline{~~-}
    & & PREFAB w/o Preview \\
    \hhline{~~-}
    & & PREFAB w/ Preview \\
    \midrule
    \multirow{5}{*}{Dependent} & \multirow{4}{*}{Quantitative Evaluation} & Self-Reported Workload \& Confidence (Figure~\ref{fig:userstudy}.a) \\
    \hhline{~~-}
    & & Self-Reported Annotation Result Quality (Figure~\ref{fig:userstudy}.c) \\
    \hhline{~~-}
    & & Temporal Characteristics of PREFAB \\
    \hhline{~~-}
    & & Annotation Consistency Between Full-Annotation and PREFAB \\
    \hhline{~--}
    & Qualitative Evaluation & Post-Interview \\
    \midrule
    \multirow{7}{*}{Control} & Game & TopDown (from AGAIN Dataset) \\
    \hhline{~--}
    & Tool & PAGAN \\
    \hhline{~--}
    & Participants & Within-Subject of 25 Participants \\
    \hhline{~--}
    & Interpolation Policy & Linear Interpolation (described in Section~\ref{subsec:interpolation}) \\
    \hhline{~--}
    & \multirow{3}{*}{Exp. Order} & Latin-Square Counterbalanced Order \\
    & & for both Annotation (Figure~\ref{fig:userstudy}.a)\\
    & & and Evaluation Sessions (Figure~\ref{fig:userstudy}.c) \\
    \bottomrule
\end{tabular}
}
\end{center}
\end{table}

\subsection{Study Design}
We conducted a user study to evaluate the effectiveness and user experience of the PREFAB method compared to traditional self-annotation. 
Table~\ref{tab:exp var} shows experiment variables of our study design.
Our study followed a within-subjects design to minimize inter-participant variability. 
The independent variable was the self-annotation method, with three conditions: Full-Annotation, PREFAB without Preview, and PREFAB with Preview. 

Here, we included a preview condition for the user study. 
For each sampled inflection region, participants first watched the corresponding clip before providing their annotations. 
We set the minimum length of an inflection region to 5 seconds, based on prior findings on reaction lag and scene comprehension (Section~\ref{subsec:inflection_detection}). 
However, because recognition and comprehension speeds vary across individuals, this minimum may not always provide sufficient context for every participant. 
For this reason, we not only compared PREFAB against full-annotation, but also examined the difference between the vanilla PREFAB method without preview and the preview-assisted PREFAB method to evaluate how additional contextual support influences annotation performance.
Providing the preview only for short clips ($\approx$ 5 seconds) would have created a mixed condition in which the presence of preview was confounded with clip length. 
Such a design would make it difficult to isolate the effect of preview itself. 
To ensure a fair and rigorous comparison, the preview was provided for all clips regardless of their duration.
All participants used the same game and annotation tool, and the independent variables were carefully controlled.
Dependent variables and control variables are detailed in the following Section~\ref{subsec:material}.

\subsection{Participants}

A total of 25 participants were recruited for this study (See details in Appendix~\ref{adx: population}). 
The participants' ages ranged from 19 to 31 (M=25.44). 
The sample consisted of 21 men and 4 women. In terms of gaming habits, participants' self-reported frequency was varied, with 6 participants playing daily, 12 weekly, 5 monthly, and 2 yearly. 
The majority (16) identified as casual gamers, while 5 identified as hardcore gamers and 4 reported rarely playing games.

Their preferred gaming platforms included PC (17), Mobile (11), and Console (3). 
Since participants could select multiple platforms, the sum of responses exceeds the total number of participants. 
The participants' favorite game genres were also diverse, with the most common being MOBA (4) and MMORPG (4).
Each participant was compensated approximately \$25 for their time.

\subsection{Materials}
\label{subsec:material}
\subsubsection{Game}
We selected TopDown, a top-down shooter game from the AGAIN dataset \cite{AGAIN}, as the experimental task. 
The shooter genre was considered suitable because it naturally evokes arousal--inducing events such as combat and item collection, while its simple control scheme minimizes learning effects across participants. 
Other candidate shooter games were excluded for specific reasons: Shootout was too simplistic, offering too few interaction events to evoke diverse affective responses, and Heist! had a steep learning curve, which could elicit frustration from skill acquisition rather than genuine emotional engagement. 
In contrast, TopDown strikes a balance between variety and simplicity, providing multiple engaging interactions without being overly complex. 
The game’s objective is to clear a maze of enemies within two minutes, which ensured a consistent and comparable experience across participants.

\subsubsection{Annotation Tool}
We used the PAGAN annotation tool \cite{PAGAN} for all conditions.
PAGAN records affective annotations in a strictly relative manner. Every annotation session begins at a neutral baseline value of 0. Participants only indicate changes in affect rather than reporting absolute affective levels.
As a result, the annotation scale has no predefined upper or lower bounds, and the recorded traces are subsequently normalized to the [0, 1] range (zero-baseline representation) per annotation session.
Accordingly, when PREFAB is used to annotate sampled clips, the resulting annotation values for each clip are also produced in this normalized 0–1 representation.

\subsubsection{Questionnaires}
We used three types of questionnaires to gather both subjective and quantitative data. First, a biographical survey, identical to the one used in the AGAIN dataset, was administered prior to the study. This information was later used as input to the PREFAB model. Second, a 5-point Likert scale survey was administered after each of the three game sessions to measure participants’ mental/physical load and confidence in their annotations (Figure~\ref{fig:userstudy}-a). Finally, after all sessions, a post-study survey was administered to measure participants’ perceived accuracy of the reconstructed annotation results, using a 5-point Likert scale, and their willingness to correct the annotation results, measured as the number of segments they wished to revise (Figure~\ref{fig:userstudy}-c).

\subsubsection{Post-Interview}
After completing all sessions and questionnaires, participants engaged in a semi-structured interview to provide qualitative insights. 
Key questions focused on their preference among the annotation methods, the primary causes of their arousal changes, and their strategy for recording changes during the annotation process. 
We also included open-ended questions about their overall experience with the PREFAB approach.

\subsubsection{Temporal Characteristics of PREFAB}
We assessed objective temporal efficiency in the user study based on the inflection regions (clips) generated by PREFAB.
Key metrics included the number of clips per session, their average and total duration, the mean time efficiency (Equation~\ref{eq:te}), and the number of ``short clips.''
We defined short clips as those lasting up to 6 seconds; this threshold incorporates a 20\% heuristic margin over the minimum clip duration (5 seconds) to account for subjective variability in perceived duration for the short clip.
These metrics were computed across all sessions, regardless of the preview condition.

\subsubsection{Annotation Consistency Between Full-Annotation and PREFAB}
\label{subsec:consistency}
We evaluated consistency using two complementary approaches: (1) the alignment of reconstructed annotation values and (2) the similarity of temporal sampling structures.
First, we assessed the quality of reconstruction. We simulated the PREFAB process on each participant's full-annotation trace---identifying inflection regions, applying zero-baselining, and interpolating---to produce a reconstructed trace.
The agreement between this reconstructed trace and the original full-annotation ground truth was quantified using three metrics: \textit{Concordance Correlation Coefficient (CCC)} \cite{ccc} for absolute agreement, \textit{Spearman's rank correlation ($\rho$)} \cite{spearman_rho} for trend similarity, and \textit{Dynamic Time Warping (DTW) Similarity} \cite{DTW} for morphological alignment.
Second, we compared the temporal characteristics of PREFAB's sampling against the inflection regions derived directly from full-annotations.
This comparison verifies whether PREFAB's selective sampling structure effectively approximates the temporal footprint of the ground-truth emotional dynamics.

\subsection{Procedure}

\begin{figure*}[t]
  \includegraphics[width=\textwidth]{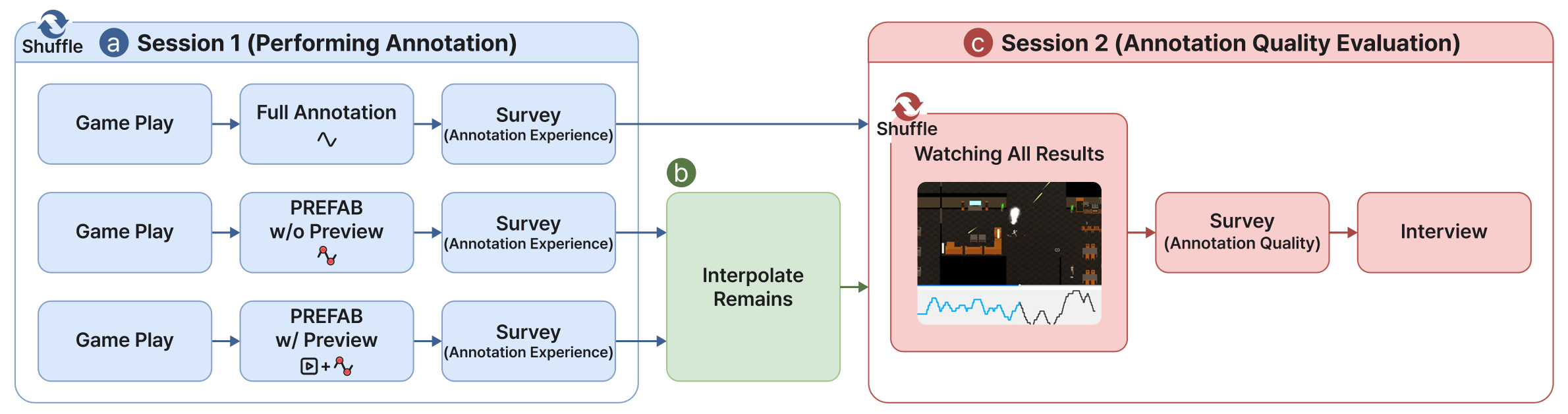}
  \caption{Overall Process of User Study}
  \Description{The user study proceeded in two sessions. In Session 1 (Performing Annotation), each participant played the game three times under different conditions: full annotation, PREFAB without preview, and PREFAB with preview. After each condition, they completed a survey on annotation experience. For PREFAB conditions, the remaining values were interpolated automatically. In Session 2 (Annotation Quality Evaluation), participants watched the results of all annotation methods in randomized order, then completed a survey on annotation quality, followed by an interview.}
  \label{fig:userstudy}
\end{figure*}

The procedure for each participant began with an explanation of the study and the acquisition of informed consent, which was approved by the [Anonymous] Institutional Review Board (IRB). 
This was followed by a tutorial on the game and the PAGAN annotation tool. 
Participants then completed a pre-study biographical survey, which was identical to the one used in the AGAIN dataset, and this biographical information was used as input to the PREFAB model.

\textbf{Performing Annotation.} Next, each participant engaged in three game sessions, with each session followed by a specific questionnaire (Figure~\ref{fig:userstudy}-a). 
The annotation method for each session was randomly assigned from the three conditions using a Latin-square counterbalance to mitigate order effects.

\textbf{Annotation Quality Evaluation.} Finally, participants completed the study by watching their gameplay videos alongside the full, reconstructed annotation results (Figure~\ref{fig:userstudy}-c). 
The partial annotations from the PREFAB methods were completed using our predefined interpolation policy (Figure~\ref{fig:userstudy}-b; see Section~\ref{subsec:interpolation}).
The videos were also presented in a randomized order, and participants were blinded to the annotation method used for each video to prevent bias. 
They then completed the final questionnaire.
After all evaluations were finished, the viewing order information was debriefed to participants, and the post-interview was conducted.

\section{Result 2: Annotation Efficiency and User Perceptions (RQ 2-4)}
\label{sec:result2}
Overall, our user study indicates that PREFAB without preview offered little benefit compared to the baseline (full-annotation). 
In contrast, providing a preview alongside the PREFAB emerged as the most effective approach--reducing mental/physical workload, improving annotation confidence, and preserving the annotation quality compared to the baseline (Figure~\ref{fig:experience}--\ref{fig:quality} and \ref{tab:annotation consistency}).
Objective analysis of the temporal characteristics revealed that the PREFAB sampled clips more finely while keeping the total duration and time-efficiency on par with the baseline (Table~\ref{tab:temporal characteristics}). This finer sampling confirms that the method effectively captured the important moments where arousal changes abruptly.
Consequently, most participants favored the PREFAB with preview condition over the other two, which were rated similarly low (Table~\ref{tab:best_worst_pref}). 
However, we found that the PREFAB conditionally mitigated temporal burden when the preview was provided, primarily evidenced by the low number of sessions where TE exceeded $0.5$ (Table~\ref{tab:temporal characteristics}).
Finally, our thematic analysis provides rich insights into how participants perceived and evaluated the PREFAB in practice, also pointing to the need for refining how previews are provided to maximize both efficiency and quality through Sections~\ref{subsec:Finding1}--\ref{subsec:finding3}.

\subsection{Quantitative Analysis}
\subsubsection{Self-Reported Responses}
We first assessed the distributional assumptions of the self-reported survey data using the Shapiro–Wilk test (see details in Table~\ref{tab:normality}). 
The results revealed that all measures violated normality ($p < 0.05$), with the exception of willingness to correct in the Baseline condition ($p = 0.141$).
Therefore, all subsequent analyses were conducted using non-parametric tests. 
We used Friedman tests to evaluate overall effects across conditions, followed by pairwise Wilcoxon signed-rank tests with Bonferroni correction for post-hoc comparisons. 
The Friedman test revealed statistically significant differences across all metrics (Mental Load: $\chi^2(2)=7.76,~p=0.021$, Physical Load: $\chi^2(2)=6.64,~p=0.036$, Confidence: $\chi^2(2)=13.58,~p=0.001$, Perceived Accuracy: $\chi^2(2)=12.18,~p=0.002$, Willingness to Correct: $\chi^2(2): 9.44,~p=0.009$), and significant pairs (see details in Table~\ref{tab:userstudy stats}) identified by the post-hoc Wilcoxon signed-rank tests are marked with asterisks in the Figures~\ref{fig:experience} and \ref{fig:quality}.

In addition, chi-square goodness-of-fit tests were employed to examine whether participants’ categorical best and worst choices of annotation methods deviated from a uniform distribution (Table~\ref{tab:best_worst_pref}). 
Results indicated that the distribution of best choices differed significantly from uniform, $\chi^2(2) = 12.08, p = 0.002$, with standardized residuals showing PREFAB with preview chosen more often than expected (+2.66) and PREFAB without preview less often (–2.19). 
In contrast, the distribution of worst choices did not significantly deviate from uniform, $\chi^2(2) = 5.36, p = 0.069$, though residuals suggested a tendency for PREFAB without preview to be more frequently regarded as worst (+1.27) and PREFAB with preview less frequently (–1.85). 

\subsubsection{Temporal Characteristics}
We additionally compared the temporal characteristics of the inflection sampling of the baseline and the PREFAB conditions (Table~\ref{tab:temporal characteristics}). 
The Shapiro-Wilk test (see details in Table~\ref{tab:normality TC}) indicated that several metrics violated the assumption of normality ($p < 0.05$), including the baseline total clip duration ($p = 0.015$) and average clip duration ($p = 0.011$). 
Therefore, non-parametric tests were performed for the statistical evaluation.
The PREFAB (aggregated across preview conditions) yielded a significantly larger number of sampled clips and a greater number of short clips ($<6$ s), while producing substantially shorter average clip durations. In contrast, the Total Clip Duration and Time Efficiency (TE) did not differ significantly from the baseline condition (Table~\ref{tab:temporal characteristics}).

\subsubsection{Annotation Consistency}
To evaluate the quality of the reconstructed annotation curves, we measured the consistency between the full-annotation traces (serving as the ground truth) and their corresponding traces reconstructed using the PREFAB sampling and interpolation methodology (Table~\ref{tab:annotation consistency}).
The results showed moderate absolute agreement (CCC = 0.67), strong ordinal correspondence (Spearman’s $\rho$ = 0.69), and high shape-level alignment (DTW similarity = 0.82).
These findings indicate that despite using only sparse inflection regions, the PREFAB can reconstruct affective trajectories that closely resemble the overall structure and relative dynamics of the full-annotations.

\subsection{Qualitative Analysis}
We followed the thematic analysis approach of Braun and Clarke \cite{braun2006} to analyze the post-interviews. Two researchers independently coded the transcripts, guided by the research questions (RQ2–RQ4), and generated initial codes. 
Through three rounds of discussion, the coding scheme was iteratively refined by adding, modifying, and removing codes. 
Based on the finalized codes, overarching themes were identified to address RQ2–RQ4, integrating quantitative results.

\subsection{Finding 1: Reduced Annotation Workload (Mental/Physical) and Partial Temporal Relief with Previews (RQ2)}
\label{subsec:Finding1}

\begin{figure}[!tb]
  \includegraphics[width=\columnwidth]{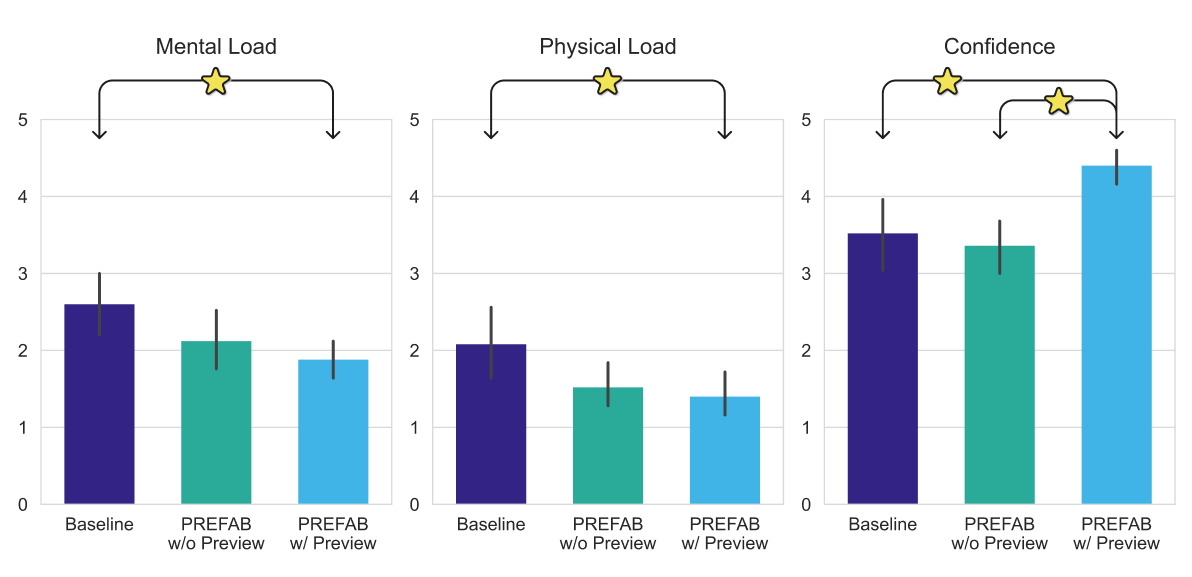}
  \caption{Results of self-reported workload and confidence during each annotation. An asterisk indicates significant difference between the pair ($p<0.05$).}
  \Description{Bar charts show self-reported mental load, physical load, and confidence across three annotation methods: baseline, PREFAB without preview, and PREFAB with preview. For mental load, baseline shows the highest average ($\approx$2.6), followed by PREFAB without preview ($\approx$2.1) and PREFAB with preview ($\approx$1.9). A significant difference is indicated between baseline and PREFAB with preview. For physical load, baseline again shows the highest average ($\approx$2.1), with lower values for PREFAB without preview ($\approx$1.6) and PREFAB with preview ($\approx$1.4). A significant difference is indicated between baseline and PREFAB with preview. For confidence, PREFAB with preview yields the highest average (≈4.4), compared to PREFAB without preview ($\approx$3.4) and baseline ($\approx$3.5). Significant differences are indicated both between baseline and PREFAB with preview, and between PREFAB without preview and PREFAB with preview. Asterisks denote statistically significant differences ($p<0.05$).}
  \label{fig:experience}
\end{figure}

Both quantitative and qualitative results indicated that PREFAB with preview reduced participants’ mental and physical burden during annotation. 
Post-hoc Wilcoxon signed-rank tests with Bonferroni correction revealed that the PREFAB with preview significantly reduced both loads compared to baseline annotation (Figure~\ref{fig:experience}).

\begin{table}[!tb]
\centering
\caption{Participants' best and worst choices of annotation method (counts), with standardized residuals in parentheses.} 
\Description{Table shows participants’ best and worst choices of annotation method, with standardized residuals in parentheses. Among 25 participants, the baseline method was selected as best by 7 (residual −0.46) and worst by 10 (residual +0.58). PREFAB without preview was chosen as best by 2 (residual −2.19) and worst by 12 (residual +1.27). PREFAB with preview was selected as best by 16 (residual +2.66) and worst by 3 (residual −1.85). These results indicate that PREFAB with preview was strongly preferred, while PREFAB without preview was most often rated worst.}
\label{tab:best_worst_pref}
\resizebox{\columnwidth}{!}{%
\begin{tabular}{c|c|c}
\toprule
Annotation Method & Best Counts (Residual) & Worst Counts (Residual) \\
\midrule
Baseline & 7 \; ($-0.46$) & 10 \; ($+0.58$) \\
PREFAB w/o preview & 2 \; ($-2.19$) & 12 \; ($+1.27$) \\
PREFAB w/ preview & 16 \; ($+2.66$) & 3 \; ($-1.85$) \\
\bottomrule
\end{tabular}
}
\end{table}

Participants’ feedback echoed these findings. 
Many reported that annotating the entire session was mentally exhausting and monotonous (P3, P4, P6, P9, P15, P22), whereas the PREFAB with preview \textit{``felt much less fatiguing''} (P1, P3, P6, P7, P9, P10, P14, P17, P20). 
Several participants emphasized that previews helped them annotate more precisely in time, reducing the effort needed to sustain attention across clips.
Many noted that having a preview allowed them to prepare mentally and annotate more accurately with less effort (P7, P10, P11, P14, P18, P19, P23, P25).
Meanwhile, participants reported that PREFAB without preview condition was burdensome, as it required immediate reactions to very short clips. 
Many noted that this increased their tension and made the annotation process more difficult (P1, P2, P7, P10, P12, P13, P16, P17, P25).

\begin{table*}[t]
\centering
\caption{Temporal characteristics of Baseline and PREFAB annotations (combining w/ and w/o preview).}
\Description{This table summarizes the temporal characteristics of the annotation methods. For each method, we report the number of detected clips, the total and average clip duration, the number of short clips (defined as clips shorter than 6 seconds), the resulting time-efficiency ratio (TE), and the count of sessions achieving high efficiency (TE $>$ 0.5). PREFAB values aggregate both preview and non-preview conditions since temporal exposure is determined solely by clip sampling. Wilcoxon signed-rank tests show that PREFAB produces significantly more clips with shorter average duration and a greater number of short clips ($p < 0.001$), while the total sampled duration and time-efficiency ratio do not differ significantly from the Baseline condition ($p = 0.19$). The session counts column for TE $>$ 0.5 is presented as descriptive statistics and was not subjected to the Wilcoxon test.}
\label{tab:temporal characteristics}
\resizebox{\textwidth}{!}{%
\begin{tabular}{c|c|c|c|c|c|c}
\toprule
\multirow{2}{*}{Method} & \multirow{2}{*}{Clip Count} & Total Clip & Average Clip & Number of & \multirow{2}{*}{Time Efficiency (TE)}& Session Counts \\
& & Duration & Duration & Short Clips ($<$ 6 s) & & of TE $>$ 0.5 \\
\midrule
Baseline (inflection sampling) & 8.00 $\pm$ 1.92 & 87.47 $\pm$ 21.40 & 11.72 $\pm$ 4.97 & 4.08 $\pm$ 2.08 & 0.28 $\pm$ 0.18 & 3 out of 25\\
PREFAB (w/ + w/o preview) & 11.44 $\pm$ 1.19 & 83.01 $\pm$ 5.31 & 7.42 $\pm$ 0.95 & 7.44 $\pm$ 1.88 & 0.32 $\pm$ 0.04 & 3 out of 50\\
\midrule
Wilcoxon p & $<$ 0.001 & 0.19 & $<$ 0.001 & $<$ 0.001 & 0.19 & N/A\\
\bottomrule
\multicolumn{7}{>{\raggedright\arraybackslash}p{\dimexpr\hsize+\tabcolsep*12\relax}}{%
    \small Session Counts of TE $>$ 0.5 when PREFAB model predicts inflection regions from the AGAIN dataset: Endless (28/112), Pirates! (22/110), Run'N'Gun (18/110), ApexSpeed (23/114), Solid (14/109), TinyCars (22/109), Heist! (16/110), Shootout (13/106), and TopDown (14/115).}
\end{tabular}
}
\end{table*}

In terms of TE, providing previews occasionally reduced efficiency rather than improving it compared to the full-annotation method. 
On average, the time spent on sampled clips using the PREFAB method accounted for approximately 68\% of the full sequence, corresponding to a mean TE ratio of 0.32 (Table~\ref{tab:temporal characteristics}). 
When the additional time required to watch previews was included, the total annotation time sometimes reached up to 136\% of the full sequence duration. 
Therefore, the total time spent could, in some cases, exceed that of the full-annotation. 
However, in sessions where the sampled clips covered less than half of the full sequence, PREFAB with preview still offered better TE than the full-annotation.

Specifically, in our user study, only 3 out of 50 sessions in the PREFAB cases had sampled regions with TE $>$ 0.5 (Table~\ref{tab:temporal characteristics}).
This means that previews had only a limited chance to support TE within our particular participant sample, even though they consistently reduced workload.
However, when we applied the same sampling process to the AGAIN dataset, which reflects a broader player population and more diverse gameplay patterns than our user study, we observed a larger number of sessions meeting this high-TE condition (Table~\ref{tab:temporal characteristics})--Endless (28/112), Pirates! (22/110), Run'N'Gun (18/110), ApexSpeed (23/114), Solid (14/109), TinyCars (22/109), Heist! (16/110), Shootout (13/106), and TopDown (14/115). 
This pattern implies that TE gains may become more common in real-world populations.

Notably, P22, who chose the PREFAB without preview as the best (Table~\ref{tab:best_worst_pref}), highlighted this trade-off, commenting that \textit{``although previews increased my confidence in annotation, the additional time cost did not justify the benefit.''} 
Interestingly, the remaining 24 participants did not explicitly perceive such temporal inefficiency; instead, they focused on the reduced workload and increased confidence afforded by the previews. 
Moreover, P22 still rated baseline annotation as the worst condition, suggesting that excessive mental fatigue from continuous annotation was perceived as more detrimental than the modest temporal inefficiency introduced by previews.

Some participants further explained that annotating the full sequence (baseline) required them to react continuously without pauses, leaving no time to organize their thoughts (P4, P9). 
This highlights the importance of providing intermittent previews to reduce cognitive strain.
However, if previews were added to every segment in the full-annotation method, the temporal efficiency would deteriorate even further, making it less efficient than PREFAB with preview. 
Thus, the combination of selective sampling and targeted previews offered by the PREFAB emerges as a necessary compromise, balancing efficiency with reduced workload and enhanced confidence.

\subsection{Finding 2: Enhanced Confidence and Effective Arousal Capture with PREFAB (RQ3)}
\label{subsec:finding2}
Participants perceived PREFAB with preview as effective in two key aspects. 
First, it enhanced their confidence during annotation. 
Second, it accurately captured the moments of arousal change. 
Quantitative analyses and interview data consistently supported both aspects.

Pairwise post-hoc Wilcoxon signed-rank tests with Bonferroni correction showed that PREFAB with preview yielded higher confidence than both baseline and PREFAB without preview (Figure~\ref{fig:experience}).
 
Interview data reinforced these results. 
Participants consistently reported that previewing a short clip before annotation gave them a stronger sense of confidence in their decisions. 
Several participants emphasized that previews increased their confidence, with one noting, \textit{``Having the preview made me feel my annotations were more accurate''} (P18). 
This view was widely shared across participants (P1, P3, P4, P6, P7, P10, P14, P15, P19, P20, P24, P25).
At the same time, some participants pointed out drawbacks of the baseline condition, explaining that watching every scene without breaks sometimes blurred their judgment, as \textit{``too many details blurred the important parts''} (P8, P11).

Participants also reflected on why PREFAB captured their affective experiences more effectively.
Most participants explained that their arousal was shaped more by situational context than by discrete game events. 
Examples of contextual factors reported by participants included encountering multiple enemies just before combat (P2, P8, P18, P24, P25), realizing that many opponents remained relative to the limited game time (P5, P14, P17), unexpectedly spotting an enemy in an unusual location (P10, P13), or facing enemies while running low on health with a health pack positioned beyond them (P7, P23).
Others described their arousal more broadly as fluctuating with contextual cues rather than single actions.
These reflections help explain why rule-based sampling, which only reacts to discrete events, exhibited limited performance in capturing participants’ arousal changes, whereas PREFAB better aligned with their lived experiences (Section~\ref{sec: technical evaluation}).

Against this backdrop, most participants felt that PREFAB effectively captured the contextual moments when their arousal either spiked or subsided (P1, P2, P6, P10, P11, P12, P13, P16, P17, P19, P20, P23). 
Notably, P5 remarked, \textit{``While annotating, I was convinced whether the clips truly represented important moments, but after reviewing the completed graphs aligned with the gameplay, I was surprised at how effective they actually were.''}

\subsection{Finding 3: PREFAB Delivers Full-Annotation Quality When Supported by Previews (RQ4)}
\begin{figure}[!tb]
  \includegraphics[width=\columnwidth]{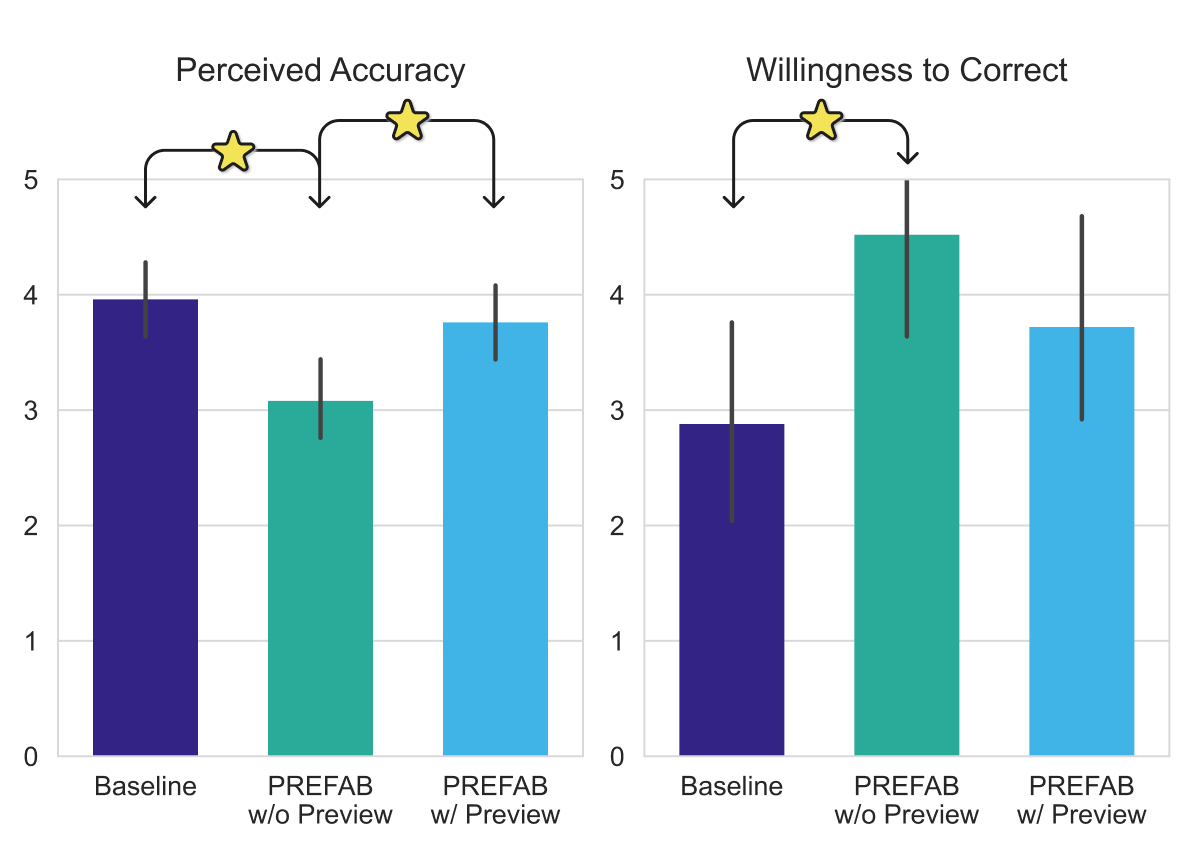}
  \caption{Results of self-reported annotation result quality evaluation. An asterisk indicates significant difference between the pair ($p<0.05$)}
  \Description{Bar charts show self-reported annotation result quality, comparing three annotation methods: baseline, PREFAB without preview, and PREFAB with preview. For perceived accuracy, baseline scored highest ($\approx$4.0), PREFAB with preview was moderately high ($\approx$3.8), and PREFAB without preview was lowest ($\approx$3.1). Significant differences are indicated between baseline and PREFAB without preview, and between PREFAB without and with preview. For willingness to correct, PREFAB without preview scored highest ($\approx$4.5), PREFAB with preview followed ($\approx$3.7), and baseline scored lowest ($\approx$2.9). A significant difference is indicated between baseline and PREFAB without preview. Asterisks indicate statistically significant differences ($p<0.05$).}
  \label{fig:quality}
\end{figure}

\begin{table}[!tb]
\centering
\caption{Average annotation consistency between full-annotation and PREFAB reconstruction.}
\Description{This table reports the average consistency between full-annotation traces and their PREFAB-based reconstructions. Consistency was assessed using three measures: the Concordance Correlation Coefficient (CCC) for absolute agreement, Spearman's rank correlation coefficient ($\rho$) for ordinal correspondence, and Dynamic Time Warping (DTW) similarity for shape alignment. Higher scores indicate stronger agreement between the reconstructed and original curves.}
\label{tab:annotation consistency}
\resizebox{0.7\columnwidth}{!}{%
\begin{tabular}{c|c|c}
\toprule
Concordance & \multirow{3}{*}{Spearman $\rho$} & \multirow{3}{*}{DTW similarity} \\
Correlation & & \\
Coefficient (CCC) & & \\
\midrule
0.67 $\pm$ 0.26 & 0.69 $\pm$ 0.23 & 0.82 $\pm$ 0.09 \\
\bottomrule
\end{tabular}
}
\end{table}

\begin{figure}[!tb]
  \includegraphics[width=\columnwidth]{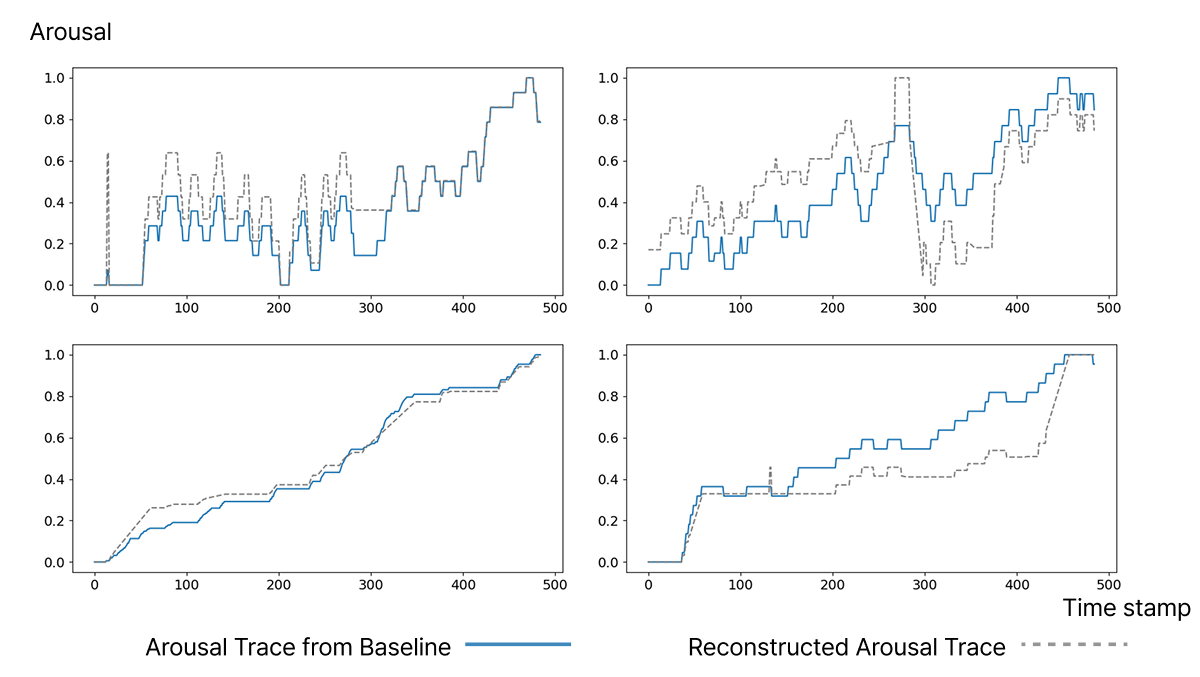}
  \caption{Comparison of the baseline arousal trace and its reconstruction via PREFAB interpolation. Despite minor deviations in scale and local simplicity, the PREFAB reconstruction (grey dashed lines) effectively captures the global trend and overall shape of the baseline arousal (blue line).}
  \Description{This figure consists of four comparative plots (2x2 grid) showing the relationship between the ground-truth Baseline Arousal Trace (solid blue line) and the PREFAB Reconstructed Arousal Trace (dashed grey line), ranging from 0.0 to 1.0 (Arousal) over 500 time units. The plots visually confirm the high consistency between the two curves.}
  \label{fig:consistency}
\end{figure}

\label{subsec:finding3}
Participants judged that PREFAB with preview delivered annotation quality comparable to the baseline. 
In particular, peaks of arousal change were captured accurately, and the interpolated segments, while somewhat simplified, were generally perceived as acceptable representations of the overall affective trajectory.

Quantitative analyses supported this perception. 
Post-hoc Wilcoxon signed-rank tests with Bonferroni correction indicated no significant differences between the baseline and PREFAB with preview in both perceived accuracy and willingness to correct (Figure~\ref{fig:quality}), suggesting that interpolation did not degrade quality compared to the full-annotation. 
In contrast, PREFAB without preview consistently received lower ratings, highlighting that the absence of previews substantially reduced perceived quality.

Objective evidence also corroborates these subjective reports.  
Table~\ref{tab:temporal characteristics} compares temporal characteristics of clips sampled by PREFAB and those derived from the baseline. 
Although the PREFAB produced a greater number of short clips and slightly shorter mean clip durations---with statistically significant differences in these metrics---the total length of sampled clips showed no significant difference across methods. 
This indicates that PREFAB's peak-detection process tended to fragment flat or neutral affective regions into smaller pieces (reflected by more short clips and shorter average duration; as illustrated in Figure~\ref{fig:reconstruction}.f), while still capturing the major peaks at comparable temporal extents to those derived from full annotation. 
Combined with PREFAB’s high F1 scores in the technical evaluation, these results suggest that the inflection regions detected by PREFAB effectively approximate the ground-truth arousal peaks despite minor over-segmentation in stable affective intervals.
This is consistent with participants' positive responses that the PREFAB captured the arousal changes well, as shown in Section~\ref{subsec:finding2}

Furthermore, reconstruction consistency analysis (Table~\ref{tab:annotation consistency}) confirms that the final interpolation based on these sampled regions remains close to full-annotation traces.  
The reconstructed arousal trajectories achieved a mean Concordance Correlation Coefficient (CCC) of 0.67 (moderate agreement), a Spearman’s $\rho$ of 0.69 (high ordinal similarity), and a DTW similarity of 0.82, indicating strong morphological alignment.  
These results collectively demonstrate that PREFAB preserves both the prominent affective peaks and the overall temporal structure of the arousal trace.
Figure~\ref{fig:consistency} compares the baseline arousal traces with those reconstructed via PREFAB interpolation from zero-represented peaks as described in Section~\ref{subsec:consistency}.

Interview data reinforced these findings. 
Most participants emphasized that the major peaks were well preserved by PREFAB, and that interpolation did not obscure the overall flow of arousal. 
As P22 responded, \textit{``Although the interpolated segments were somewhat linear and less nuanced, the main trajectory was maintained''}. 
Others similarly commented that \textit{``the interpolated parts looked too uniform''} (P7, P11, P16, P17, P21), but emphasized that the final outcome did not differ substantially from what they really felt. 
Several participants even expressed surprise at how effective the interpolated outputs were when viewed alongside the completed graphs. 
\textit{``I was surprised at how effective it actually was''} (P5, P9). 

At the same time, participants suggested potential improvements, such as \textit{``allowing an additional round of correction after interpolation''} (P18) or \textit{``making the rationale for clip selection more transparent''} (P9).

These findings indicate that using PREFAB alone, without previews, may lower the perceived quality of final annotations due to insufficient temporal context for the short clips. 
In contrast, when previews were provided, participants reported no perceived difference in annotation quality compared to full-annotation. 
This highlights that the provision of previews is critical for maintaining annotation fidelity. 

\section{Discussion}
\subsection{Trade-Off between Workload Reduction and Temporal Efficiency}
Our mixed-method analysis (Section~\ref{sec:result2}) shows that PREFAB with preview achieves self-annotation quality comparable to full annotation, while significantly lowering subjective workload. 
This reduction provides annotators with greater capacity to sustain longer and larger tasks. 
However, the benefit of reduced workload comes with the cost of additional time spent watching previews. 
As shown in our results (Section~\ref{subsec:Finding1}), when the cumulative length of sampled clips exceeds about half of the original task duration, the added preview time can outweigh the time saved, potentially surpassing the duration of full annotation.

The decision to adopt a 5-second minimum clip length was motivated by prior findings on cognitive lag in self-annotation (a delay of approximately 1–3s between perceiving and labeling affect, occasionally up to 6s) \cite{time_lag1, time_lag2, time_lag3} and evidence that viewers generally require at least 3 s of exposure to grasp the context of a scene \cite{clip_duration}. 
While this design ensured that clips were perceivable, our findings highlight that perceivability and comfort in annotation are distinct dimensions. 
Participants in the no-preview condition consistently reported that reacting to very short clips was \textit{``stressful''} and \textit{``fatiguing''}. 
On average, in the user study, PREFAB sampled an average of 11.44 clips per task, and more than half of them (7.44) were shorter than 6 seconds which is close to minimum clip length (Table~\ref{tab:temporal characteristics}). 
Such a predominance of short clips likely increased participants’ sense of fatigue when no preview was provided.

These findings indicate that PREFAB with preview is an effective method for reducing workload, but temporal efficiency remains a challenge. 
Addressing this trade-off requires more selective and adaptive preview strategies. 
One promising direction is to use recent foundation VLMs \cite{VLM} to produce short descriptions of sampled clips, so that participants can access context without spending extra time watching them.
Another complementary approach is to establish a threshold for ``too short'' clips based on empirical measures of annotator fatigue, and provide previews only for those segments.
These approaches could maintain the workload benefits of PREFAB while improving its temporal efficiency, but they remain conceptual suggestions; implementing and validating them will require further research.

\subsection{Triggers of Arousal Change Beyond Events: Opportunities with Unobtrusive Modalities}
Many participants reported that situational factors within the game context—such as encountering multiple enemies just before combat, running low on health with a health pack positioned beyond opponents, or realizing that limited time remained with many enemies still present—had a stronger influence on their arousal than discrete event-driven cues like a score change or a single combat action (Section~\ref{subsec:finding2}). 
This aligns with our performance results, where the rule-based sampling method that relied only on game logs achieved lower F1 scores compared to modeling approaches such as PREFAB or regression (Section~\ref{sec: technical evaluation}).

Although deep neural networks are effective at capturing patterns from digitized task observations, such observations do not contain direct information about users’ internal states. 
Our findings suggest that relying solely on these features is insufficient to capture contextual influences. 
While PREFAB achieved the highest F1 score among all methods, its performance still remained below 0.7 (Figure~\ref{fig:performance}), leaving considerable room for improvement.

These findings point to the need for advanced techniques that enable PREFAB to better recognize the situational context of the main task. 
One potential direction is to incorporate physiological signals beyond task-driven features. 
Prior work has demonstrated the value of combining visual and auditory cues for modeling players’ internal states, suggesting that multimodal information can provide richer context \cite{Pixels_and_Sounds}. 
In this regard, studies such as OCEAN \cite{OCEAN} and PresUP \cite{presup} have explored the use of physiological sensors to detect opportune moments for intervention. 
However, these approaches often rely on relatively heavy sensing devices, which may limit their practicality in the main task by interfering with the user experience.

As a more practical alternative, lightweight textile sensors could be explored \cite{Yiyue1, Yiyue2}. 
For example, in the gaming context, by embedding textile sensors into the controller parts that players naturally touch, it may be possible to infer affective states from touch patterns without disrupting playability \cite{AffectiveTouchSurvey}. 
Integrating such unobtrusive sensing modalities could allow PREFAB to access contextual signals that go beyond raw task features, ultimately improving its ability to model arousal in real-world gaming environments.

\subsection{Limitations}
This study has several limitations that open up directions for future work. 
First, our user study was conducted only with the TopDown game from the AGAIN dataset, and the participant pool was heavily skewed toward male and gamer populations. 
While this controlled setting allowed us to carefully evaluate PREFAB in a consistent and engaging context, it also limits the generalizability of our findings to broader affective computing scenarios.
In particular, gender and gaming experience may influence affective response patterns or annotation behaviors, potentially biasing the resulting models.
Future work should therefore examine effectiveness of PREFAB in more diverse populations and in other domains such as film, education, or training, where arousal is influenced by more diverse contextual and demographic factors.

Second, PREFAB requires initial annotated data for training. 
This process inevitably involves collecting full-annotation of complete sessions from participants. 
In large-scale projects where annotations accumulate over time, our method could become increasingly efficient. 
However, for smaller projects, the upfront cost of data collection may outweigh its benefits. 
Addressing this challenge will require strategies such as transfer learning and cross-domain adaptation, for example leveraging PREFAB models trained on one domain to bootstrap annotation in another \cite{AffectFAL}.

Third, our evaluation did not examine generalization performance of PREFAB model beyond the training conditions. 
PREFAB model was only tested on the game and affective state (arousal) used for training. 
It remains unclear how well the method extends to unseen environments or to other affective dimensions such as valence or dominance. 
Addressing this will be important for establishing broader robustness.

Finally, our interpolation policy was restricted to a simple linear strategy that extended the momentum of the most recent annotated region into the unannotated segments (Section~\ref{subsec:interpolation}). 
While most participants judged the overall quality of the interpolated annotations to be satisfactory, some reported that the linearly interpolated parts felt unnatural (e.g., P7, P11, P16, P17, P21). 
A heuristic inspection of full-annotation data from the user study suggests that users’ annotation behaviors may follow several distinct patterns, for example, monotonically increasing, maintaining linear slopes after inflections, remaining flat, or varying gradients of rise and fall. 
Although our study did not investigate these patterns in depth, future work could analyze such individual annotation styles and develop personalized interpolation policies that better capture users’ affective trajectories.

\section{Conclusion}
This paper introduced PREFAB, a novel approach to reducing the cognitive and temporal cost of retrospective affect self-annotation. Grounded in the Peak-End Rule and the ordinal nature of emotion, PREFAB employs preference-learning techniques to train models that predict affective inflection points, focusing annotation on key moments rather than requiring full annotation. Through model evaluation, PREFAB consistently outperformed baseline methods across nine games, and a user study validated that PREFAB with preview maintained annotation quality while lowering participants’ workload and increasing their confidence. 
These results suggest that focusing on key emotional moments is sufficient to preserve the overall affective trajectory. Nevertheless, our findings highlight a trade-off: while previews effectively reduce cognitive burden, temporal efficiency (TE) improvements are conditional, primarily evidenced by the sparse number of sessions achieving high efficiency (TE $>$ 0.5).
Future work should investigate adaptive preview strategies and richer contextual cues beyond game features to further enhance performance. In addition, our interpolation policy was limited to a simple linear approach; developing more sophisticated, potentially personalized interpolation strategies could better capture diverse annotation patterns. Overall, PREFAB offers a practical and human-centered solution for scalable affective data collection, reducing cost without sacrificing quality.

\begin{acks}
We appreciate the high-performance GPU computing support of HPC-AI Open Infrastructure via GIST SCENT. This work was supported by Institute of Information \& communications Technology Planning \& Evaluation (IITP) grant funded by the Korea government (MSIT) (No.2019-0-01842, Artificial Intelligence Graduate School Program (GIST)). This work was supported by the National Research Foundation of Korea(NRF) grant funded by the Korea government(MSIT) (RS-2025-16902996).
\end{acks}

\bibliographystyle{ACM-Reference-Format}
\bibliography{reference}

\appendix
\clearpage
\section{Model Architecture Ablation Study}
\label{adx:ablation}

We ablated two design choices in PREFAB, (1) FiLM-based conditioning on biography information and (2) an auxiliary classification task, to quantify their individual and joint contributions. 
The ablation was conducted on the TopDown game, which is also the target domain of the user study. 
We compare four model variants: Use-Nothing, Biography-only (FiLM), Auxiliary-only, and All (PREFAB).
Performance metrics follows Section~\ref{sec: technical evaluation}: F1 score, TE, and $\Delta$TE.

Table~\ref{tab:ablation} shows the result of the ablation study.
A one-way ANOVA was run separately for F1, TE, and $\Delta$TE with four conditions as the factor, and it shows strong statistical significances for all metrics (F1: $F(3, 8811) = 132.6$, $p<0.001$, TE: $F(3, 8811) = 886.75$, $p<0.001$, $\Delta$TE: $F(3, 8811) = 364.06$, $p<0.001$).
Bonferroni-corrected post-hoc pairwise t-tests indicated that all pairs differed significantly for all metrics (all $p<0.001$). 
When we use all techniques into the PREFAB model, it attains the the F1 and TE (including $\Delta$TE).
In case of biography-only (FiLM method), it slightly raises F1 over Use-Nothing case but exhibits very low TE, suggesting an over-fragmented reconstruction (many inflection regions), which is inefficient and misaligned with the TE of the ground truth.
Conversely, Auxiliary task-only reduces F1 relative to Use-Nothing but yields higher TE than Use-Nothing and Biography-only, indicating a smoother reconstruction with fewer inflection derivations.

\begin{figure*}[t]
\renewcommand{\thefigure}{A1}
  \includegraphics[width=\textwidth]{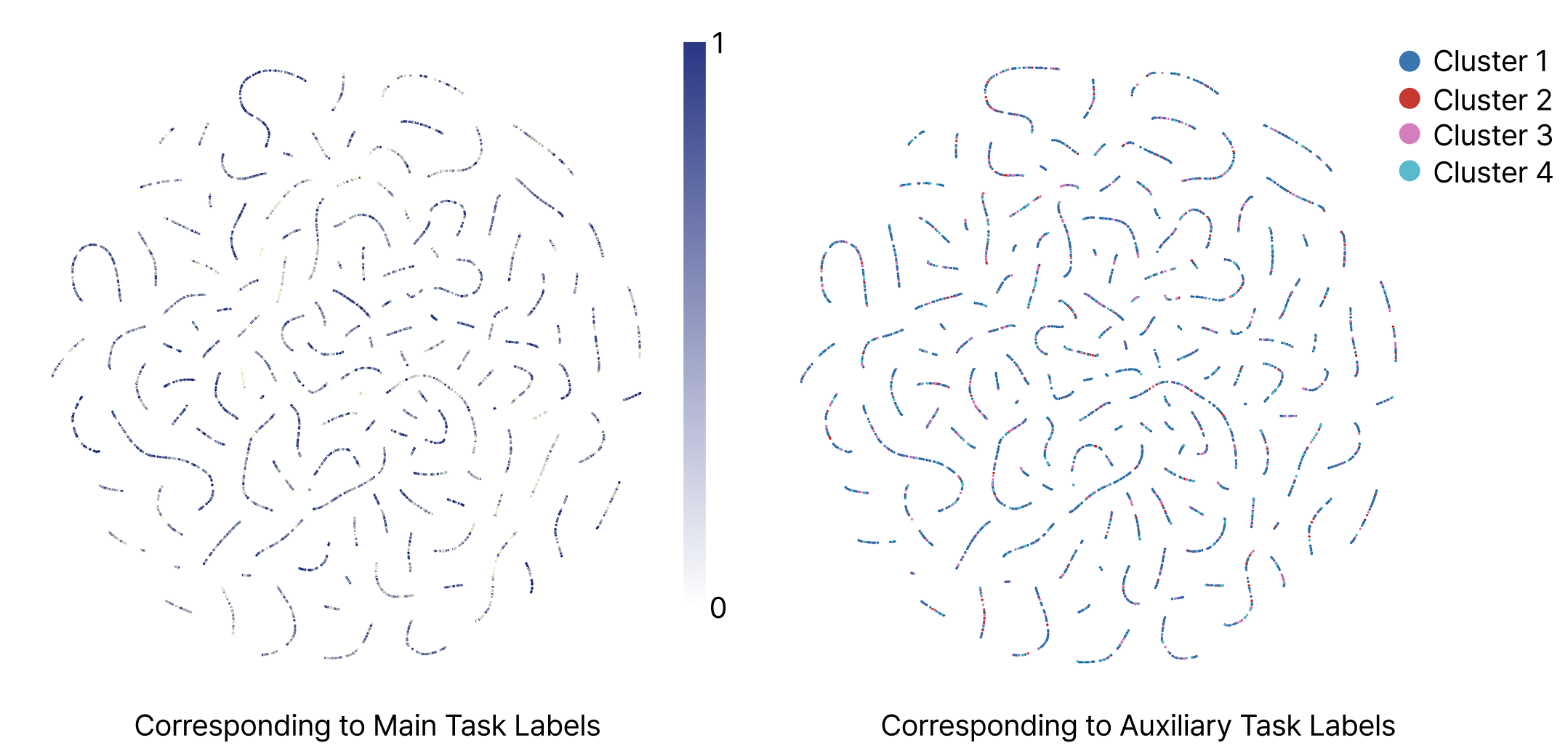}
  \caption{t-SNE results for the `use-nothing' case. While clusters are clearly separated, features within each cluster are intermixed across labels for both the main task (left) and the auxiliary task (right).}
  \Description{Figure shows t-SNE visualizations for the Use-Nothing condition. On the left is the main task embedding, and on the right is the auxiliary task embedding. In both plots, several clusters appear visually distinct, but labels within each cluster are intermixed across categories. This suggests that although blobs are well separated in the t-SNE plane, they do not align with the true label structure.}
  \label{fig:nothing tsne}
\end{figure*}

\begin{figure*}[t]
\renewcommand{\thefigure}{A2}
  \includegraphics[width=\textwidth]{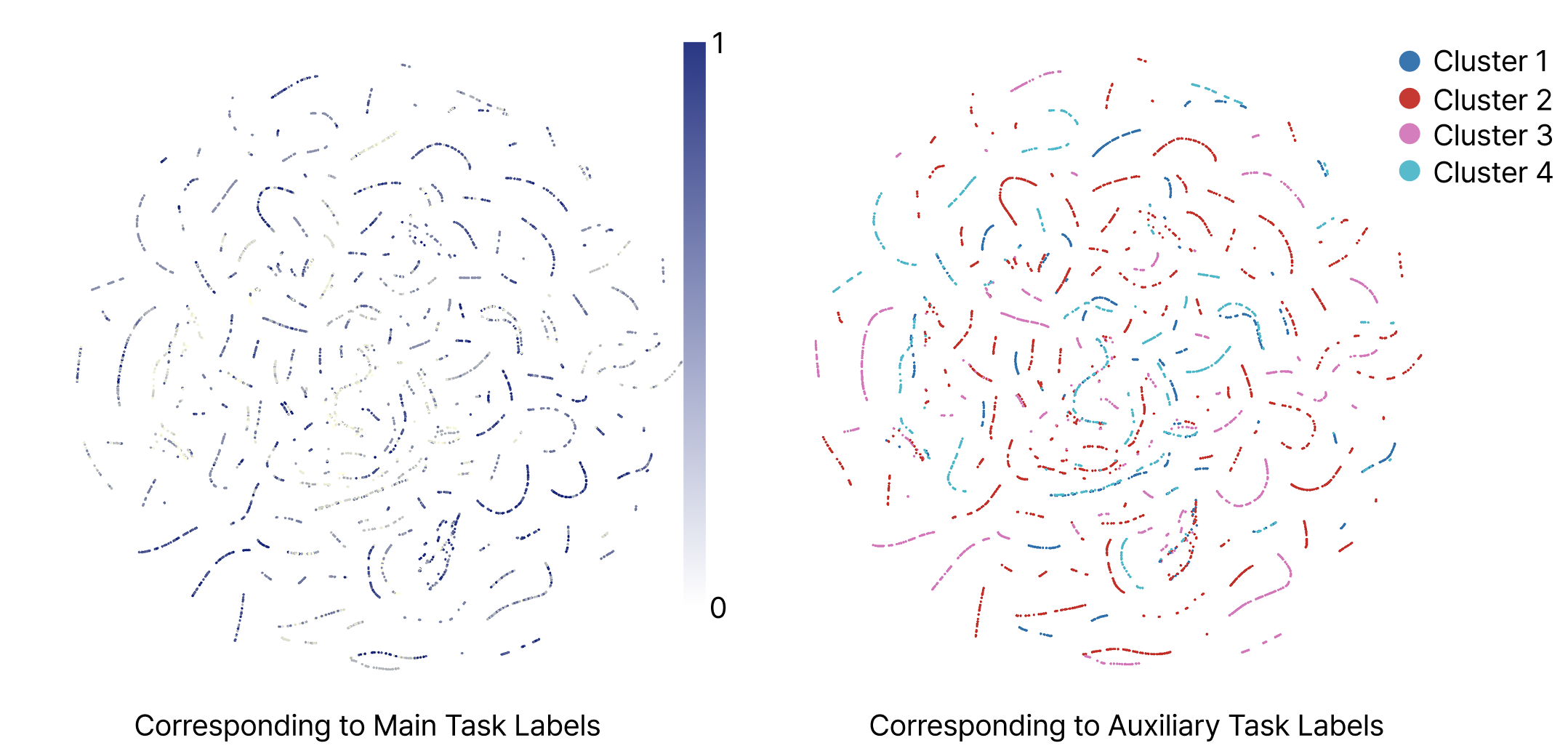}
  \caption{t-SNE results for the `biography information only (FiLM)' case. Features are well separated across labels for both the main task (left) and auxiliary task (right), though cluster boundaries are the noisiest among the four cases.}
  \Description{Figure shows t-SNE visualizations for the Biography information only (FiLM) condition. The left plot is for the main task and the right for the auxiliary task. Compared to the Use-Nothing case, features are better aligned with labels, and label cohesion within clusters is visibly improved, as similar labels co-locate more tightly. However, cluster boundaries are noisier and less distinct than in other cases, indicating stronger intra-cluster alignment but weaker global separability.}
  \label{fig:bio only tsne}
\end{figure*}

\begin{figure*}[t]
\renewcommand{\thefigure}{A3}
  \includegraphics[width=\textwidth]{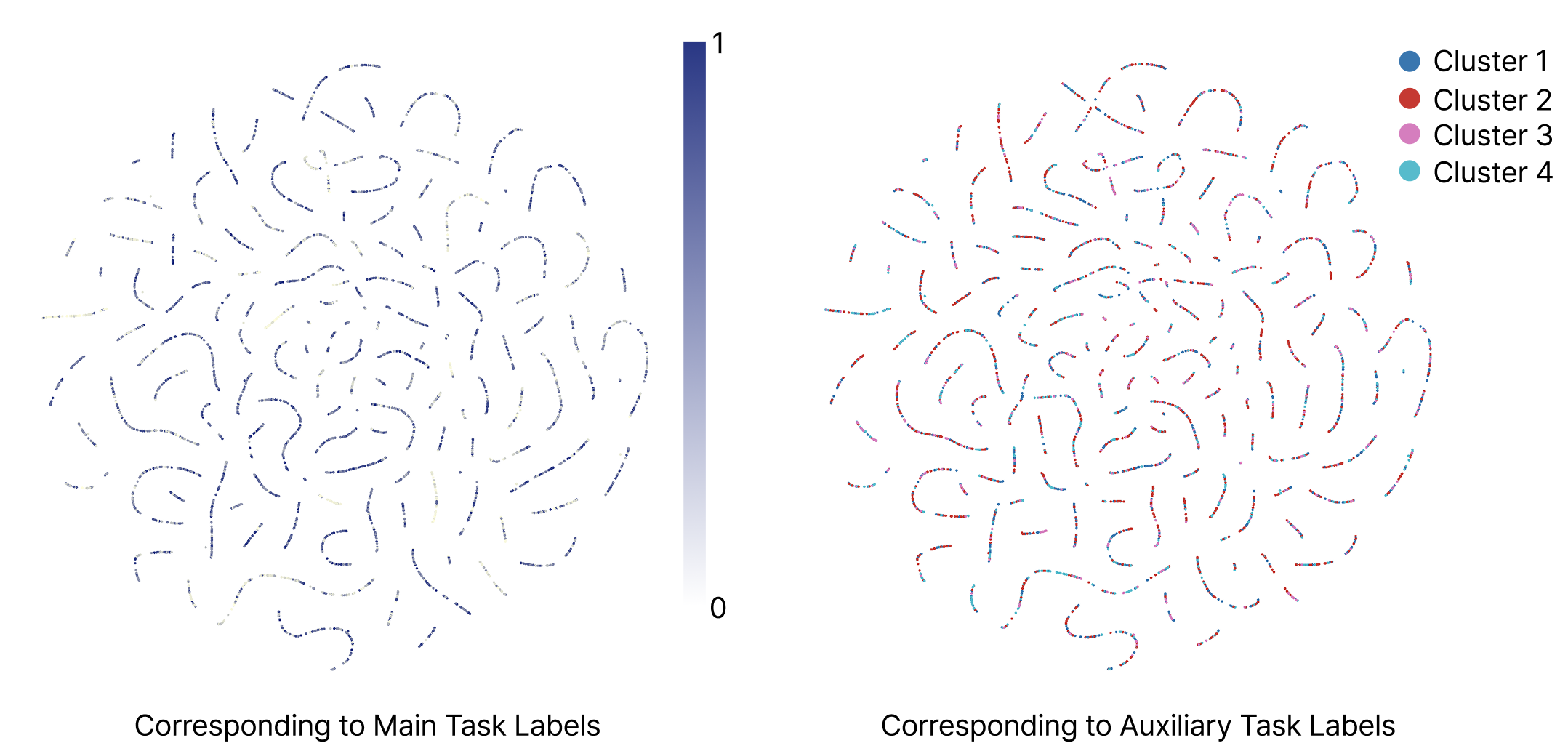}
  \caption{t-SNE results for the Auxiliary Classifier only case. Clusters are clearly separated, but features within each cluster are intermixed across labels for both the main task (left) and the auxiliary task (right).}
  \Description{Figure shows t-SNE results for the Auxiliary Classifier only condition. The left panel is for the main task and the right for the auxiliary task. Clusters are more sharply separated at a global scale, with clearer coarse partitions than in the Use-Nothing or FiLM-only cases. However, within each cluster, label mixing remains prominent, and labels are not cleanly separated. This indicates that the auxiliary objective mainly regularizes the overall geometry, improving time efficiency by smoothing reconstructions, but does not by itself resolve fine-grained label alignment.}
  \label{fig:aux only tsne}
\end{figure*}

\begin{figure*}[t]
\renewcommand{\thefigure}{A4}
  \includegraphics[width=\textwidth]{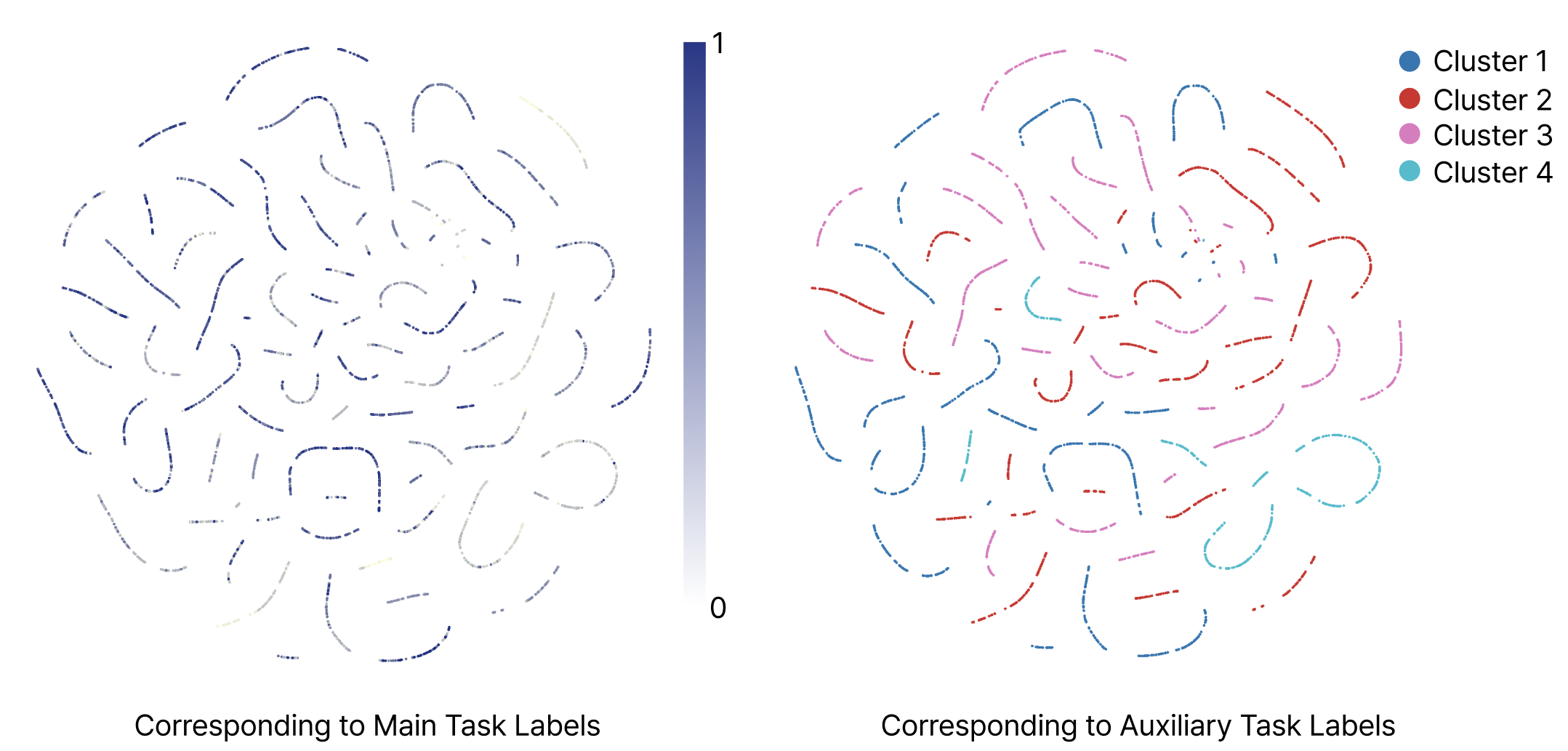}
  \caption{t-SNE results for the PREFAB case. Clusters are clearly separated, and features within each cluster are distinctly divided for both the main task (left) and the auxiliary task (right).}
  \Description{Figure shows t-SNE results for the PREFAB condition. The left panel corresponds to the main task and the right to the auxiliary task. Compared to other variants, the PREFAB embedding achieves both clear inter-cluster separation and high label purity within clusters. The clusters are fewer in number, more compact, and more coherently aligned with both labeling schemes, reflecting a concise latent geometry. These visual patterns correspond to the quantitative results in which PREFAB achieved the highest F1, the highest time efficiency, and the lowest $\Delta$TE.}
  \label{fig:PREFAB tsne}
\end{figure*}

Qualitative differences are also evident in the latent-space projections. 
Figures~\ref{fig:nothing tsne}–\ref{fig:PREFAB tsne} visualize 2-D t-SNE projections of latent vectors, color-coded by the main task labels (continuous relative arousal; left) and the auxiliary task labels (trend clusters; right).
In the Use-Nothing condition (Figure~\ref{fig:nothing tsne}), several visually separated blobs emerge in the t-SNE plane; however, labels remain heavily intermixed within each blob for both the main task (left) and the auxiliary task (right). 
This indicates that local neighborhoods in the embedding do not correspond cleanly to label structure, implying fragile decision behavior under small perturbations.

With Biography-only (FiLM; Figure~\ref{fig:bio only tsne}), conditioning realigns the representation so that label-wise cohesion within clusters improves; labels of similar value tend to co-locate more tightly. 
At the same time, inter-cluster boundaries appear noisier and less well delimited, suggesting that FiLM primarily enhances intra-cluster label alignment while weakening global separability in the geometry.

The Auxiliary-only variant (Figure~\ref{fig:aux only tsne}) shows the complementary pattern: global partitions in the latent geometry look sharper, and clusters are more cleanly delineated at a coarse scale, yet label mixing persists within clusters for both main and auxiliary tasks. 
Consequently, we refrain from interpreting this as improved inter-class separability with respect to labels; rather, the auxiliary objective mainly regularizes global geometry without resolving label intermixing on its own. 
This picture is consistent with the table-level behavior in which Auxiliary-only yields higher TE (smoother reconstructions with fewer spurious inflection regions) relative to Use-Nothing and biography-only conditions.

When All (PREFAB) techniques are combined (Figure~\ref{fig:PREFAB tsne}), the embedding exhibits both properties simultaneously: inter-cluster boundaries are clear, and intra-cluster label purity is high for the main and auxiliary tasks. 
Visually, the number of coherent clusters is smaller and each cluster is more compact, indicating a concise latent geometry aligned with both labeling schemes. 
This joint improvement mirrors the quantitative outcome where PREFAB attains the highest F1, the highest TE, and the lowest $\Delta$TE among the four variants.

Note that because t-SNE is a nonlinear projection for visualizing latent space, we use it solely as a qualitative aid and base our conclusions on the statistically significant differences reported in Table~\ref{tab:ablation}.

\begin{table*}[t]
\renewcommand{\thetable}{A1}
\centering
\caption{Comparison of F1 Scores, Time Efficiency (TE), and $\Delta$TE (average gap between ground-truth and predicted TE) in the ablation study. $\uparrow$ indicates higher values are better and $\downarrow$ indicates lower values are better.} 
\Description{Table A1 reports ablation results. The full PREFAB model achieved the best performance, with the highest F1 and time efficiency and the lowest $\Delta$TE, while all ablated variants showed significantly worse results (ANOVA, p < 0.001).}
\label{tab:ablation}
\resizebox{\textwidth}{!}{%
\begin{tabular}{c|c|c|c|c|c|c}
\toprule
Model & F1 Score $\uparrow$ & Time Efficiency $\uparrow$ & $\Delta$TE $\downarrow$ & F1 ANOVA & TE ANOVA & $\Delta$TE ANOVA\\
\midrule
Use Nothing & 0.568 $\pm$ 0.137 & 0.182 $\pm$ 0.045 & 0.210 $\pm$ 0.160 & &&\\
\hhline{----~~~}
Biography information Only (FiLM) & 0.572 $\pm$ 0.134 & 0.043 $\pm$ 0.025 & 0.325 $\pm$ 0.193 & F(3,~8811) = 132.60 &F(3,~8811)=886.75&F(3,~8811)=364.06\\
\hhline{----~~~}
Auxiliary Classifier Only & 0.517 $\pm$ 0.132 & 0.215 $\pm$ 0.050 & 0.189 $\pm$ 0.129 &, p<0.001, $\eta_p^2 = 0.043$ &, p<0.001, $\eta_p^2 = 0.232$ &, p<0.001, $\eta_p^2 = 0.110$ \\
\hhline{----~~~}
All (PREFAB) & 0.663 $\pm$ 0.125 & 0.442 $\pm$ 0.135 & 0.124 $\pm$ 0.082 &&&\\
\bottomrule
\end{tabular}
}
\end{table*}

\FloatBarrier

\clearpage
\section{Feature Selection for The Event-Driven Rule-Based Sampling}
\label{adx:feature selection}
We implement a na\"ive heuristic baseline for event-driven sampling by exploiting correlations between changes in game logs and changes in arousal in the AGAIN dataset. 
The dataset provides around 120 per-frame game features, including 28 common features used in all games and additional game-specific features. 
To capture relationship between arousal changes and in-game events, we compute frame-to-frame differences for each feature $x_t$ and the annotated arousal $y_t$:
$\Delta x_t = x_t - x_{t-1}$ and $\Delta y_t = y_t - y_{t-1}$. We then compute Pearson’s correlation $r(\Delta x, \Delta y)$ per game.

Across all nine games, after excluding trivial time indices (e.g., timestamp), the change in \textit{score} shows the highest absolute correlation with the change in arousal (Table~\ref{tab:feature selection}). 
Based on this observation, we hypothesize that intervals with nonzero \textit{score} changes correspond to affective inflection regions, and we define a rule-based sampler accordingly. 
First, we segment the timeline into continuous intervals during which \textit{score} changes continue. 
If the duration of a segment is $\leq 5$ seconds (the minimum inflection region duration as defined in Section~\ref{subsec:inflection_detection}), we treat it as a short event and select its midpoint as the inflection point. 
Otherwise, we treat it as a long event and select both the segment start and end as inflection points. 
This yields a simple, reproducible baseline that requires no model training while aligning the sampling strategy with the strongest change-level correlate of arousal in the dataset.

\begin{table*}[t]
\renewcommand{\thetable}{B1}
\centering
\caption{Top-5 absolute Pearson correlations between feature changes and arousal changes, by game.}
\Description{Table B1 lists the top-5 absolute Pearson correlations between feature changes and arousal changes for each game. For ApexSpeed, the strongest correlations are with score (0.419), player_lap (0.377), and player_gas_pedal (0.345). For Endless, score (0.364), player_delta_distance (0.304), and player_movement (0.304) show the highest correlations. For Heist!, score (0.472) dominates, followed by weaker correlations such as player_health (−0.131). For Pirates!, score (0.329) is the top feature, while others show much weaker correlations. For Run’N’Gun, score (0.465) and player_health (−0.242) stand out. For Shootout, multiple bot-related features strongly correlate, including score (0.537), bot_diversity (0.407), and bot_reloading (0.402). For Solid, the top features are score (0.443), player_lap (0.375), and player_standing (0.287). For TinyCars, score (0.351), player_lap (0.254), and player_standing (0.215) are strongest. For Topdown, score (0.543) is highest, followed by player_health (−0.249). Across all games, score consistently shows the highest correlation with arousal changes, while other features vary by game.}
\label{tab:feature selection}
\resizebox{\textwidth}{!}{%
\begin{tabular}{c|c|c|c|c|c}
\toprule
Game & 1st & 2nd & 3rd & 4th & 5th \\
\midrule
ApexSpeed   & score (0.419) & player\_lap (0.377) & player\_gas\_pedal (0.345) & player\_standing (0.220) & bot\_distance\_to\_way\_point (-0.212) \\
Endless     & score (0.364) & player\_delta\_distance (0.304) & player\_movement (0.304) & player\_speed\_x (0.298) & bot\_movement (0.084) \\
Heist!      & score (0.472) & player\_health (-0.131) & player\_crouching (0.065) & player\_shooting (0.063) & event\_intensity (0.063) \\
Pirates!    & score (0.329) & player\_has\_powerup (0.085) & player\_health (-0.046) & object\_intensity (0.042) & pick\_ups\_visible (0.042) \\
Run'N'Gun   & score (0.465) & player\_health (-0.242) & visible\_bot\_count (0.125) & bot\_diversity (0.112) & bot\_health (0.089) \\
Shootout    & score (0.537) & bot\_diversity (0.407) & bot\_reloading (0.402) & bot\_health (0.397) & visible\_bot\_count (0.333) \\
Solid       & score (0.443) & player\_lap (0.375) & player\_standing (0.287) & player\_speed (0.168) & bot\_distance\_to\_way\_point (-0.119) \\
TinyCars    & score (0.351) & player\_lap (0.254) & player\_standing (0.215) & visible\_bot\_count (-0.151) & player\_speed (0.148) \\
Topdown     & score (0.543) & player\_health (-0.249) & destrucible\_count (0.159) & player\_shooting (0.084) & player\_projectile\_count (0.084) \\
\bottomrule
\end{tabular}
}
\end{table*}

\FloatBarrier

\clearpage
\section{Detailed Statistics}
\label{adx: stats}

\begin{table*}[b]
\begin{center}
\renewcommand{\thetable}{C1}
  \caption{F1 and $\Delta$TE results for each game ($\uparrow$: higher is better, $\downarrow$: lower is better). All comparisons showed statistically significant differences ($p < .001$) with large effect sizes ($\eta_p^2 > 0.14$)}
  \Description{The table reports F1 scores and $\Delta$TE values for nine games (Apex, Endless, Heist!, Pirates!, Run'N'Gun, Shootout, Solid, TinyCars, TopDown) across five annotation strategies: Random, Uniform, Rule-Based, Regression, and PREFAB. For every game, PREFAB yields the highest F1 score, ranging roughly from 0.63 to 0.69, substantially outperforming all baseline methods. Uniform sampling often produces the second-highest F1 but shows large $\Delta$TE values (approximately 0.28–0.38), indicating inefficient region selection that covers broad portions of the sequence. Rule-Based methods frequently exhibit high $\Delta$TE and low F1, while Regression provides moderate performance on both metrics. Random sampling yields mid-range F1 scores (about 0.48–0.52) and $\Delta$TE values (around 0.15–0.16). PREFAB consistently achieves the smallest $\Delta$TE in all games except Endless, where it ranks second. ANOVA tests indicate significant differences among methods for both F1 and $\Delta$TE across all games ($p < .001$).}
  \label{tab:performance}
  \resizebox{0.9\textwidth}{!}{%
  \begin{tabular}{c|c|c|c|c|c}
    \toprule
    Game & Method & F1 $\uparrow$ & $\Delta$TE $\downarrow$ & F1 ANOVA & $\Delta$TE ANOVA\\
    \midrule
    \multirow{5}{*}{Apex} & Random & 0.502 $\pm$ 0.106 & 0.155 $\pm$ 0.129 & \multirow{5}{*}{$F(4, 55183)=2346.29,~p<0.001$} & \multirow{5}{*}{$F(4,55183)=2267.10,~p<0.001$}\\
    & Uniform & 0.611 $\pm$ 0.114 & 0.312 $\pm$ 0.187 & &\\
    & Rule-Based & 0.557 $\pm$ 0.111 & 0.232 $\pm$ 0.184 & &\\
    & Regression & 0.575 $\pm$ 0.146 & 0.211 $\pm$ 0.162 & &\\
    & PREFAB & 0.648 $\pm$ 0.130 & 0.134 $\pm$ 0.082 & &\\
    \midrule
    \multirow{5}{*}{Endless} & Random & 0.457 $\pm$ 0.117 & 0.154 $\pm$ 0.117 & \multirow{5}{*}{$F(4,53115)=3091.62,~p<0.001$} & \multirow{5}{*}{$F(4,53115)=3243.97,~p<0.001$}\\
    & Uniform & 0.567 $\pm$ 0.106 & 0.381 $\pm$ 0.181 & &\\
    & Rule-Based & 0.553 $\pm$ 0.113 & 0.288 $\pm$ 0.176  & &\\
    & Regression & 0.519 $\pm$ 0.121 & 0.276 $\pm$ 0.176  & &\\
    & PREFAB & 0.635 $\pm$ 0.141 & 0.224 $\pm$ 0.156 & & \\
    \midrule
    \multirow{5}{*}{Heist!} & Random & 0.500 $\pm$ 0.098 & 0.149 $\pm$ 0.117 & \multirow{5}{*}{$F(4,51433)=12083.37,~p<0.001$} & \multirow{5}{*}{$F(4,51433)=3117.30,~p<0.001$} \\
    & Uniform & 0.590 $\pm$ 0.087 & 0.294 $\pm$ 0.157 & & \\
    & Rule-Based & 0.331 $\pm$ 0.131 & 0.294 $\pm$ 0.185  & &\\
    & Regression & 0.561 $\pm$ 0.130 & 0.206 $\pm$ 0.158  & &\\
    & PREFAB & 0.637 $\pm$ 0.117 & 0.132 $\pm$ 0.088 & & \\
    \midrule
    \multirow{5}{*}{Pirates!} & Random & 0.506 $\pm$ 0.114 & 0.155 $\pm$ 0.110 & \multirow{5}{*}{$F(4, 51433)=11619.92,~p<0.001$} & \multirow{5}{*}{$F(4,51433)=2499.81,~p<0.001$} \\
    & Uniform & 0.622 $\pm$ 0.106 & 0.311 $\pm$ 0.182 & & \\
    & Rule-Based & 0.359 $\pm$ 0.126 & 0.249 $\pm$ 0.183  & &\\
    & Regression & 0.602 $\pm$ 0.136 & 0.176 $\pm$ 0.161  & &\\
    & PREFAB & 0.685 $\pm$ 0.130 & 0.126 $\pm$ 0.107 & & \\
    \midrule
    \multirow{5}{*}{Run'N'Gun} & Random & 0.505 $\pm$ 0.109 & 0.152 $\pm$ 0.125 & \multirow{5}{*}{$F(4,51433)=9388.87,~p<0.001$} & \multirow{5}{*}{$F(4,51433)=2115.66,~p<0.001$} \\
    & Uniform & 0.610 $\pm$ 0.103 & 0.287 $\pm$ 0.181 & & \\
    & Rule-Based & 0.377 $\pm$ 0.134 & 0.235 $\pm$ 0.151 & & \\
    & Regression & 0.552 $\pm$ 0.131 & 0.188 $\pm$ 0.149 & & \\
    & PREFAB & 0.681 $\pm$ 0.121 & 0.120 $\pm$ 0.096 & & \\
    \midrule
    \multirow{5}{*}{Shootout} & Random & 0.525 $\pm$ 0.127 & 0.163 $\pm$ 0.112 & \multirow{5}{*}{$F(4,47481)=3881.53,~p<0.001$} & \multirow{5}{*}{$F(4,47481)=1794.84,~p<0.001$} \\
    & Uniform & 0.632 $\pm$ 0.121 & 0.277 $\pm$ 0.188 & & \\
    & Rule-Based & 0.489 $\pm$ 0.107 & 0.172 $\pm$ 0.129  & &\\
    & Regression & 0.539 $\pm$ 0.141 & 0.173 $\pm$ 0.132  & &\\
    & PREFAB & 0.656 $\pm$ 0.142 & 0.112 $\pm$ 0.085 & & \\
    \midrule
    \multirow{5}{*}{Solid} & Random & 0.514 $\pm$ 0.087 & 0.158 $\pm$ 0.128 & \multirow{5}{*}{$F(4,50430)=4558.01,~p<0.001$} & \multirow{5}{*}{$F(4,50430)=2228.82,~p<0.001$}\\
    & Uniform & 0.596 $\pm$ 0.089 & 0.283 $\pm$ 0.168 & & \\
    & Rule-Based & 0.451 $\pm$ 0.120 & 0.172 $\pm$ 0.139 & & \\
    & Regression & 0.507 $\pm$ 0.109 & 0.138 $\pm$ 0.109 & & \\
    & PREFAB & 0.632 $\pm$ 0.133 & 0.134 $\pm$ 0.089 & & \\
    \midrule
    \multirow{5}{*}{TinyCars} & Random & 0.496 $\pm$ 0.105 & 0.154 $\pm$ 0.125 & \multirow{5}{*}{$F(4,50430)=2254.46,~p<0.001$} & \multirow{5}{*}{$F(4,50430)=3032.10,~p<0.001$}\\
    & Uniform & 0.591 $\pm$ 0.100 & 0.337 $\pm$ 0.180 & & \\
    & Rule-Based & 0.507 $\pm$ 0.137 & 0.210 $\pm$ 0.161 & & \\
    & Regression & 0.561 $\pm$ 0.137 & 0.239 $\pm$ 0.166 & & \\
    & PREFAB & 0.634 $\pm$ 0.127 & 0.135 $\pm$ 0.079 & & \\
    \midrule
    \multirow{5}{*}{TopDown} & Random & 0.505 $\pm$ 0.091 & 0.149 $\pm$ 0.118 & \multirow{5}{*}{$F(4,56232)=14929.97,~p<0.001$} & \multirow{5}{*}{$F(4,56232)=3884.72,~p<0.001$} \\
    & Uniform & 0.597 $\pm$ 0.094 & 0.295 $\pm$ 0.172 & & \\
    & Rule-Based & 0.331 $\pm$ 0.129 & 0.313 $\pm$ 0.191 & & \\
    & Regression & 0.582 $\pm$ 0.128 & 0.170 $\pm$ 0.142 & & \\
    & PREFAB & 0.663 $\pm$ 0.125 & 0.124 $\pm$ 0.082 & & \\
    \bottomrule
\end{tabular}
}
\end{center}
\end{table*}

\begin{table*}[b]
\centering
\renewcommand{\thetable}{C2}
\caption{Results of the Shapiro-Wilk test for the user study ($n = 25$).}
\Description{Table reports the results of the Shapiro–Wilk test for the user study (n=25). For each condition (Baseline, PREFAB without preview, and PREFAB with preview), the table shows the W statistic and corresponding p-value for the five measures: mental load, physical load, confidence, perceived accuracy, and willingness to correct. 
Across most measures and conditions, p-values fall below the .05 threshold—often below .001—indicating that normality is violated in the majority of cases. These results justify the use of non-parametric tests (Friedman and Wilcoxon signed-rank tests) 
for the main statistical analyses.}
\label{tab:normality}
\resizebox{0.9\textwidth}{!}{%
\begin{tabular}{c|c|c|c|c|c|c}
\toprule
Method & Component & Mental Load & Physical Load & Confidence & Perceived Accuracy & Willingness to Connect \\
\midrule
\multirow{2}{*}{Baseline} & W statistic & 0.909 & 0.779 & 0.873 & 0.843 & 0.939 \\
& p & 0.03 & $<$ 0.001& 0.005& 0.001& 0.141\\
\midrule
\multirow{2}{*}{PREFAB w/o preview} & W statistic & 0.845 & 0.708 & 0.836 & 0.862 & 0.889 \\
& p & 0.001& $<$ 0.001& $<$ 0.001& 0.003 & $<$ 0.011\\
\midrule
\multirow{2}{*}{PREFAB w/ preview} & W statistic & 0.796 & 0.593 & 0.731 & 0.864 & 0.872 \\
& p & $<$ 0.001& $<$ 0.001& $<$ 0.001& 0.003& 0.004\\
\bottomrule
\end{tabular}
}
\end{table*}

\begin{table*}[b]
\centering
\renewcommand{\thetable}{C3}
\caption{Numerical results (means) and post-hoc pairwise Wilcoxon signed-rank tests with Bonferroni correction ($\alpha = 0.0167$) for Figures~\ref{fig:experience} and~\ref{fig:quality} ($n = 25$).}
\Description{Table shows numerical means and post-hoc pairwise Wilcoxon signed-rank tests with Bonferroni correction for five measures: mental load, physical load, confidence, perceived accuracy, and willingness to correct, comparing baseline, PREFAB without preview, and PREFAB with preview. Mean scores: Baseline — mental load 2.60, physical load 2.08, confidence 3.52, perceived accuracy 3.96, willingness to correct 2.88. PREFAB without preview — 2.12, 1.52, 3.36, 3.08, 4.52. PREFAB with preview — 1.88, 1.40, 4.40, 3.76, 3.72. Significant results: Baseline vs PREFAB without preview shows lower mental load (p=0.016), lower physical load (p=0.043), lower perceived accuracy (p=0.014), and higher willingness to correct (p=0.012) for PREFAB without preview. Baseline vs PREFAB with preview shows lower mental load (p=0.003), lower physical load (p=0.011), and higher confidence (p=0.005) for PREFAB with preview. PREFAB without vs with preview shows significantly higher confidence (p<0.001) and higher perceived accuracy (p=0.003) for PREFAB with preview.}
\label{tab:userstudy stats}
\resizebox{0.8\textwidth}{!}{%
\begin{tabular}{c|c|c|c|c|c}
\toprule
Method & Mental Load & Physical Load & Confidence & Perceived Accuracy & Willingness to Correct \\
\midrule
Baseline (1) & 2.60 & 2.08 & 3.52 & 3.96 & 2.88\\
PREFAB w/o preview (2) & 2.12 & 1.52 & 3.36 & 3.08 & 4.52 \\
PREFAB w/ preview (3) & 1.88 & 1.40 & 4.40 & 3.76 & 3.72 \\
\midrule
(1) vs (2) & p=0.053 & p=0.043 & p=0.544 & p=0.014 & p=0.012\\
(1) vs (3) & p=0.003 & p=0.011 & p=0.005 & p=0.386 & p=0.121 \\
(2) vs (3) & p=0.196 & p=0.367 & p$<$0.001 & p=0.003 & p=0.073 \\
\bottomrule
\end{tabular}
}
\end{table*}

\begin{table*}[b]
\centering
\renewcommand{\thetable}{C4}
\caption{Results of the Shapiro-Wilk test for the temporal characteristics.}
\Description{This table presents the results of the Shapiro-Wilk test for normality on the temporal characteristics data (Clip Count, Total Clip Duration, Average Clip Duration, Number of Short Clips, and TE) for both the Baseline and PREFAB methods. Each method displays the W statistic and the corresponding p-value. For the Baseline condition, Total Clip Duration ($p=0.015$) and Average Clip Duration ($p=0.011$) show statistically significant violation of normality ($p < 0.05$). All other metrics for the Baseline and all metrics for the aggregated PREFAB (w/ + w/o preview) condition indicate that the null hypothesis of normality cannot be rejected ($p > 0.05$).}
\label{tab:normality TC}
\resizebox{0.8\textwidth}{!}{%
\begin{tabular}{c|c|c|c|c|c|c}
\toprule
\multirow{2}{*}{Method}& \multirow{2}{*}{Component}& \multirow{2}{*}{Clip Count} & Total Clip & Average Clip & Number of & \multirow{2}{*}{TE} \\
&&&Duration&Duration&Short Clips ($<$ 6s) & \\
\midrule
\multirow{2}{*}{Baseline} & W statistic & 0.975 & 0.897 & 0.889 & 0.936 &0.975\\
& p & 0.78 & 0.015 & 0.011 & 0.12 & 0.78\\
\midrule
\multirow{2}{*}{PREFAB (w/ + w/o preview)} & W statistic & 0.976 & 0.973 & 0.973 & 0.959 & 0.976 \\
& p & 0.79 & 0.72 & 0.72 & 0.39 & 0.79\\
\bottomrule
\end{tabular}
}
\end{table*}

\FloatBarrier

\clearpage
\section{User Study Population}
\label{adx: population}

\begin{table*}[b]
\begin{center}
\renewcommand{\thetable}{D1}
  \caption{Participant biographical information. The format follows that of the AGAIN dataset.}
  \Description{Table summarizes the biographical information of 25 participants, following the AGAIN dataset format. Each row lists a participant with attributes: age, gender, play frequency, gamer type, favorite game platform, and favorite game genre. Ages range from 19 to 31. The sample includes 4 females and 21 males. Play frequency varies from daily to yearly, with most reporting weekly play. Gamer types are primarily casual, with several hardcore and a few rarely playing. Preferred platforms are mostly PC and mobile, with some console users. Favorite genres span tactical FPS, RTS, MOBA, MMORPG, RPG, survival, rhythm, puzzle, and sports simulations.}
  \label{tab:bio}
  \resizebox{0.9\textwidth}{!}{%
  \begin{tabular}{c|c|c|c|c|c|c}
    \toprule
    Participant & Age & Gender & Play Frequency & Gamer Type & Favorite Game Platform & Favorite Game Genre \\
    \midrule
    P1 & 25 & Male & Weekly & Casual & PC & Tactical FPS \\
    P2 & 19 & Male & Weekly & Casual & PC & RTS \\
    P3 & 24 & Male & Monthly & Casual & PC & MOBA \\
    P4 & 27 & Male & Weekly & Hardcore & PC & Soccer Sim \\
    P5 & 26 & Male & Monthly & Rarely Playing & PC & Tactical FPS \\
    P6 & 23 & Female & Daily & Casual & Mobile & Trainer RPG \\
    P7 & 26 & Male & Weekly & Casual & PC & MOBA \\
    P8 & 30 & Male & Monthly & Rarely Playing & Console & Trainer RPG \\
    P9 & 26 & Male & Weekly & Casual & PC & MMORPG \\
    P10 & 30 & Female & Yearly & Rarely Playing & PC & Tactical FPS \\
    P11 & 22 & Male & Daily & Casual & PC, Mobile & Survival \\
    P12 & 23 & Male & Daily & Hardcore & PC, Mobile & MOBA \\
    P13 & 21 & Male & Monthly & Casual & Mobile & Metroidvania \\
    P14 & 26 & Male & Weekly & Casual & Mobile & MOBA \\
    P15 & 25 & Female & Monthly & Casual & Mobile & Stacking Puzzle \\
    P16 & 27 & Male & Daily & Casual & PC, Mobile & Soccer Sim \\
    P17 & 29 & Male & Weekly & Casual & Mobile & MMORPG \\
    P18 & 27 & Male & Weekly & Hardcore & Console & Action RPG \\
    P19 & 19 & Male & Daily & Hardcore & PC, Mobile & FPS \\
    P20 & 29 & Male & Weekly & Casual & PC & MMORPG \\
    P21 & 27 & Male & Weekly & Casual & PC & MMORPG \\
    P22 & 31 & Male & Weekly & Casual & PC & Card Battle \\
    P23 & 26 & Male & Weekly & Casual & PC, Mobile & RTS \\
    P24 & 24 & Male & Daily & Hardcore & PC, Console & Rhythm \\
    P25 & 24 & Female & Yearly & Rarely Playing & Mobile & Stacking Puzzle \\
    \bottomrule
\end{tabular}
}
\end{center}
\end{table*}

\end{document}